\documentclass[10pt,twocolumn,letterpaper]{article}

\usepackage[pagenumbers]{cvpr} 

\newcommand{\tick}{\color{Green}{\ding{52}}}
\newcommand{\cross}{\color{red}{\ding{56}}}

\usepackage{comment}

\definecolor{cvprblue}{rgb}{0.21,0.49,0.74}
\usepackage[pagebackref,breaklinks,colorlinks,allcolors=cvprblue]{hyperref}

\usepackage{amsmath}
\usepackage{graphicx}
\usepackage{placeins}
\usepackage{bm}
\usepackage{colortbl}
\usepackage{multirow}
\usepackage{pifont}
\usepackage[table]{xcolor}
\usepackage[accsupp]{axessibility}

\definecolor{green1st}{RGB}{100,190,100}
\definecolor{green2nd}{RGB}{190,210,180}
\definecolor{green3rd}{RGB}{210,240,210}

\usepackage{arydshln}
\def\paperID{****}
\def\confName{CVPR}
\def\confYear{2026}

\newcommand{\method}{OVI-MAP\xspace}

\title{\method: Open-Vocabulary Instance-Semantic Mapping}

\author{
Zilong Deng$^{1,3}$, Federico Tombari$^{2,4}$, Marc Pollefeys$^{1,5}$, Johanna Wald$^{2,\dagger}$, Daniel Barath$^{1,2,\dagger}$\\
\tt\small{$^\dagger$Equal contribution}\\
$^{1}$ETH Zurich, $^{2}$Google, $^{3}$University of Zurich, $^{4}$TU Munich, $^{5}$Microsoft\\
}

\begin{document}
\maketitle

\begin{abstract}
Incremental open-vocabulary 3D instance-semantic mapping is essential for autonomous agents operating in complex everyday environments. However, it remains challenging due to the need for robust instance segmentation, real-time processing, and flexible open-set reasoning. Existing methods often rely on the closed-set assumption or dense per-pixel language fusion, which limits scalability and temporal consistency. 
We introduce OVI-MAP that decouples instance reconstruction from semantic inference. We propose to build a class-agnostic 3D instance map that is incrementally constructed from RGB-D input, while semantic features are extracted only from a small set of automatically selected views using Vision-Language Models. 
This design enables stable instance tracking and zero-shot semantic labeling throughout online exploration. 
Our system operates in real time and outperforms state-of-the-art open-vocabulary mapping baselines on standard benchmarks. Code available at \hyperlink{https://ovi-map.github.io}{\textit{https://ovi-map.github.io}}.

%
%
%
%
\end{abstract}    
\section{Introduction}
\label{sec:intro}

Semantic and instance 3D mapping are foundational capabilities for embodied perception, supporting downstream tasks like language-conditioned navigation, manipulation and facilitating queryable scene understanding for both AR/VR and robotics~\cite{chen2022nlmap, huang2022vlmap, jatavallabhula2023conceptfusion, takmaz2025search3d, laina2025findanything}. 
In indoor environments, voxel-based scene representations -- most commonly Truncated Signed Distance Fields (TSDFs) -- have become a standard choice due to their real-time fusion, robustness to drift, and ability to maintain dense and consistent geometry for planning and interaction~\cite{newcombe2011kinectfusion, oleynikova2017voxblox, grinvald2019voxblox++}. 
Recent systems further couple volumetric reconstruction with semantics, yielding panoptic maps that are temporally consistent and easy to query~\cite{yu2024panopticrecon, yu2025panopticrecon++, weder2024labelmaker, miao2024volumetric, yamazaki2024openfusion}.

However, existing pipelines are predominantly \emph{closed-set}: they assume a fixed semantic ontology, learn class-dependent predictors, and store integer class labels per representation unit (\eg, voxel, point). 
Extending these designs to \emph{open-set} recognition is non-trivial for several reasons. 
First, open-set features extracted from Vision-Language Models (VLMs) are high-dimensional and continuous; naively storing them at voxel-level resolution leads to substantial computational and memory overhead. 

Second, recent volumetric mapping systems~\cite{yamazaki2024openfusion, miao2024volumetric} rely on semantic labels to guide instance segmentation and association. Without semantics, object instance grouping becomes unstable and prone to fragmentation. 
Third, without consistent 3D instances, aggregating per-pixel open-set features over time is noisy due to occlusions, viewpoint changes, background noise and inconsistent 2D segmentations. 
While recent segmentation models, such as SAM~\cite{kirillov2023sam}, provide high-quality object proposals, running them per frame is computationally expensive and therefore unsuitable for real-time online mapping. 

\begin{figure}[t]
    \centering
    \footnotesize
    \includegraphics[width=\linewidth]{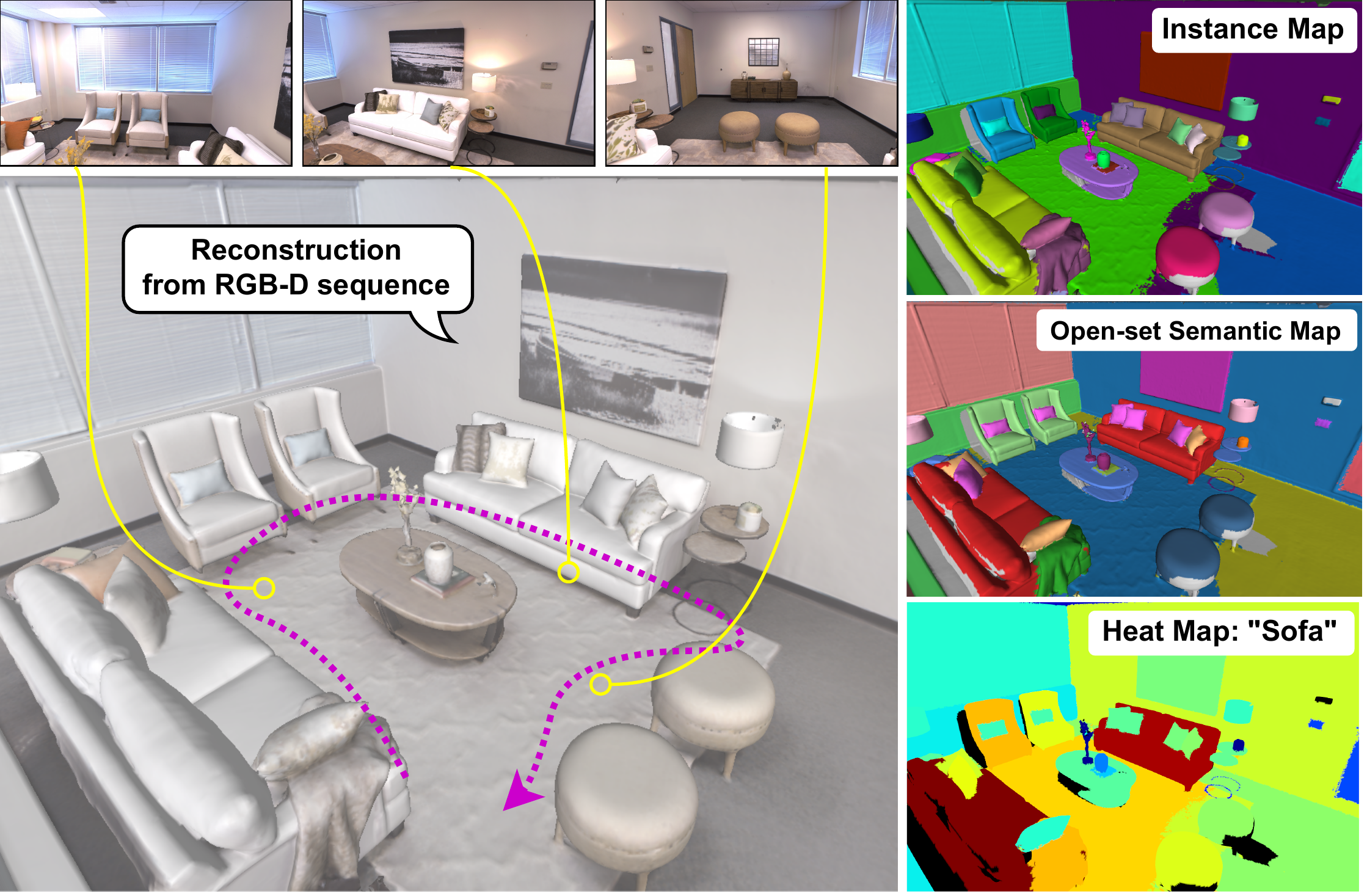}
    \caption{
        \textbf{Overview of OVI-MAP.} Given a streaming RGB-D sequence with camera poses, OVI-MAP incrementally reconstructs a volumetric 3D scene while maintaining a \emph{class-agnostic} instance map. Semantic information is then assigned in a \emph{zero-shot} manner using selectively chosen views, enabling open-set object recognition. Our method supports real-time, open-world scene reconstruction with instance-level semantic understanding.
    }
    \label{fig:task_definition}
    \vspace{-2mm}
\end{figure}

To address these limitations, we propose a pipeline that decouples instance mapping from open-set semantic recognition. 
Our approach first performs multi-view \emph{class-agnostic} instance segmentation and incrementally lifts these instances into a global TSDF-based representation, producing a consistent \emph{instance map} across frames without requiring semantic labels. 
On top of this instance map, we introduce an \emph{object-centric view selection strategy} that identifies informative viewpoints for semantic extraction. 
Semantic features are queried from VLMs only when a new viewpoint provides non-redundant coverage of the 3D object's surface, improving both efficiency and robustness while avoiding per-voxel storage of high-dimensional embeddings.

\noindent
In summary, our contributions are:
\begin{itemize}
    \item Class-agnostic online instance reconstruction pipeline that maintains a consistent 3D instance map independent of semantic categories.
    \item Object-centric incremental view selection mechanism that adaptively selects frames for feature extraction, reducing redundant VLM queries.
    \item An efficient, open-set semantic aggregation procedure that associates image regions with 3D instances and enables robust zero-shot semantic recognition.
\end{itemize}
We show on the ScanNet~\cite{dai2017scannet} and Replica~\cite{straub2019replica} datasets that our proposed decoupled design leads to state-of-the-art instance mapping and open-set semantic segmentation while maintaining online processing rates.

\begin{table}[t]
\centering
\resizebox{1.0\linewidth}{!}{
\begin{tabular}{l|c c c c}
\hline
\textbf{Method} & \textbf{3D Recon.} & \textbf{Inst. Map} & \textbf{Zero-Shot Sem.} & \textbf{Online} \\
\hline
\textbf{Ours} 
    & TSDF & \tick & \tick & \tick \\
OVO-SLAM~\cite{martins2024ovoslam} 
    & PCL/3DGS & \tick & \tick & \tick \\
OpenFusion~\cite{yamazaki2024openfusion} 
    & TSDF & \cross & \cross & \tick \\
ConceptFusion~\cite{jatavallabhula2023conceptfusion} 
    & PCL & \cross & \tick & \tick \\
OV-Octree-Graph~\cite{wang2025open} 
    & PCL & \tick & \tick & \cross \\
DCSEG~\cite{wiedmann2025dcseg} 
    & 3DGS & \tick & \cross & \cross \\
OpenNeRF~\cite{engelmann2024opennerf} 
    & NeRF~\cite{mildenhall2021nerf} & \cross & \tick & \cross \\
Semantic Gauss~\cite{guo2024semantic} 
    & 3DGS & \cross & \tick & \cross \\
OpenMask3D~\cite{takmaz2023openmask3d} 
    & \cross & \tick & \tick & \cross \\
OpenScene~\cite{peng2023openscene} 
    & \cross & \cross & \tick & \cross \\
Mask3D~\cite{schult2022mask3d} 
    & \cross & \tick & \cross & \cross \\
\hline
\end{tabular}}
\caption{
State-of-the-art online and offline semantic/instance mapping methods. PCL stands for Point Cloud, and 3DGS denotes 3D Gaussian Splatting~\cite{kerbl20233dgs}.}
\label{tab:method_compare}
\vspace{-2mm}
\end{table}

\section{Related Work}
\label{sec:related}

\textbf{Open-Vocabulary Image Understanding} has advanced rapidly with Vision-Language Models (VLMs), like CLIP~\cite{radford2021clip} and SigLIP~\cite{zhai2023sigmoid}, that align visual and textual embeddings through large-scale contrastive pretraining. 
To enable pixel-level reasoning, many works produce dense or region-level semantic features. LSeg~\cite{li2022lseg} predicts dense CLIP-aligned embeddings. 
OpenSeg~\cite{ghiasi2022openseg} and OVSeg~\cite{liang2023ovseg} refine segmentation using region proposals. 
SAM~\cite{kirillov2023sam} and its derivatives~\cite{xiong2024efficientsam, ravi2024sam2} offer category-agnostic masks, but produce over-segmented results. 
Works like Mask2Former~\cite{li2024semanticsam, cheng2022mask2former} provide semantic grounding only in closed-set, and X-Decoder~\cite{zou2023xdecoder, zou2023seem} provides a unified architecture for open-vocabulary segmentation. 
While these works show promising results on 2D images, they do not natively scale to 3D and there do not provide long-term, consistent segmentation across different views. We leverage these 2D models but focus on \emph{incremental 3D} scenes and separate instance mapping from semantic recognition.

\vspace{1mm} \noindent
\textbf{3D Semantic Understanding with VLMs.}  
A growing class of methods lifts 2D open-vocabulary features into 3D by fusing multi-view semantic cues into volumetric, point-cloud, or neural field representations.  
OpenScene~\cite{peng2023openscene}, PLA~\cite{ding2023pla}, and ConceptFusion~\cite{jatavallabhula2023conceptfusion} distill pixel-aligned VLM embeddings into point-level representations for open-vocabulary reasoning.  
Methods like CLIP‑Fields~\cite{shafiullah2022clipfields}, SemAbs~\cite{ha2022semantic}, 3DSS‑VLG~\cite{xu20243dssvlg}, and PoVo~\cite{mei2025povo} propagate CLIP-aligned features across views and align them with text embeddings.  
More recently, Gaussian splatting and NeRF approaches~\cite{kerr2023lerf, engelmann2024opennerf, qin2024langsplat,piekenbrinck2025opensplat3d,yu2024panopticrecon,zhi2021semantic-nerf,guo2024semantic,li2025scenesplat} process 3D scenes by directly embedding semantics into 3D Gaussian primitives or neural implicit fields.  
However, these methods often require global scene optimization or dense 3D semantic fields, making them sub-optimal for online, incremental mapping.  
In contrast, our method avoids this limitation and computes semantics only \emph{per-instance} via selective view aggregation, yielding scalable real-time mapping.

\vspace{1mm} \noindent
\textbf{3D Instance Segmentation and Panoptic Mapping}
is commonly performed directly on point clouds using architectures such as PointNet++~\cite{qi2017pointnet++}, MinkowskiNet~\cite{choy2019minkowski}, and Mask3D~\cite{schult2022mask3d}. 
These models, however, rely on large-scale annotated 3D datasets and are trained in closed-set settings. 
To reduce reliance on manual labels, recent works~\cite{yang2023sam3d,huang2024segment3d,ye2024gaussiangrouping,yin2024sai3d} propagate 2D mask priors into 3D, but still require frequent segmentation updates and do not target online operation.
Incremental panoptic mapping in indoor scenes has been explored using TSDF-based fusion~\cite{grinvald2019voxblox++, li2020incremental, miao2024volumetric, yamazaki2024openfusion, zheng2024mapadapt}, yet these systems couple instance association with closed-set semantic label prediction~\cite{he2017maskrcnn,cheng2022mask2former,zou2023seem}, making them incompatible with open-set recognition.

\vspace{1mm} \noindent
\textbf{Open-Set Instance-Level Semantic Mapping.}
OpenMask3D~\cite{takmaz2023openmask3d}, Search3D~\cite{takmaz2025search3d} extracts CLIP-like features for 3D instances from multi-view crops, demonstrating that \emph{instance-level} semantic aggregation is more stable than pixel-level fusion~\cite{peng2023openscene}. 
However, these works assume instance masks are available and do not address online mapping. 
Works like O3D-SIM~\cite{nanwani2024o3dsim} and DCSEG~\cite{wiedmann2025dcseg} combine clustering-based instance segmentation and class-dependent semantic masks from 2D VLMs, achieving offline instance-semantic segmentation with pre-defined semantic labels.
Recent, OVO-SLAM~\cite{martins2024ovoslam} and OpenGS-SLAM~\cite{yang2025opengsslam} combine SLAM with open-vocabulary instance queries, but they still rely on time-consuming and over-segmented SAM~\cite{kirillov2023sam} outputs and do not reason about view selection efficiency. OpenVox~\cite{deng2025openvox} maintains the voxel-based instance map while its semantic codebook relies on Yolo-world~\cite{cheng2024yoloworld} with pre-set label promptings.

\begin{figure*}[t]
    \centering
    \begin{subfigure}{0.85\linewidth}
        \includegraphics[width=\linewidth]{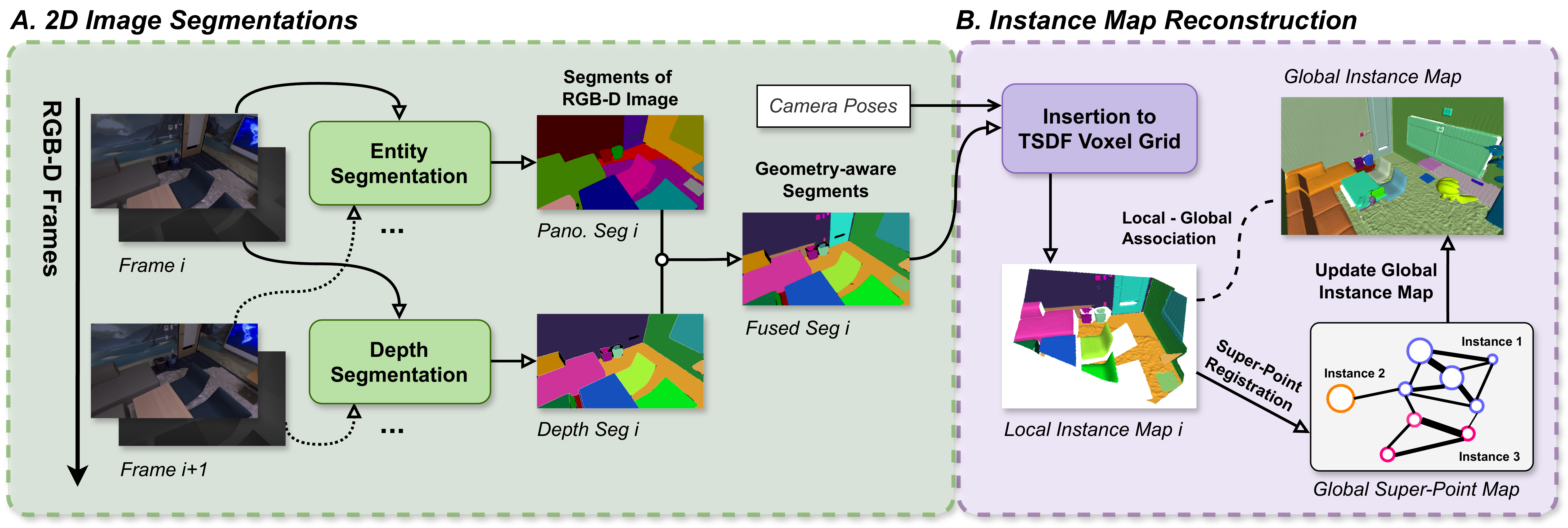}
        \caption{Class-Agnostic Instance Map Reconstruction.}
    \label{fig:pipeline1}
    \end{subfigure}
    \vspace{0.2em}
    \centering
    \begin{subfigure}{0.88\linewidth}
        \includegraphics[width=\linewidth]{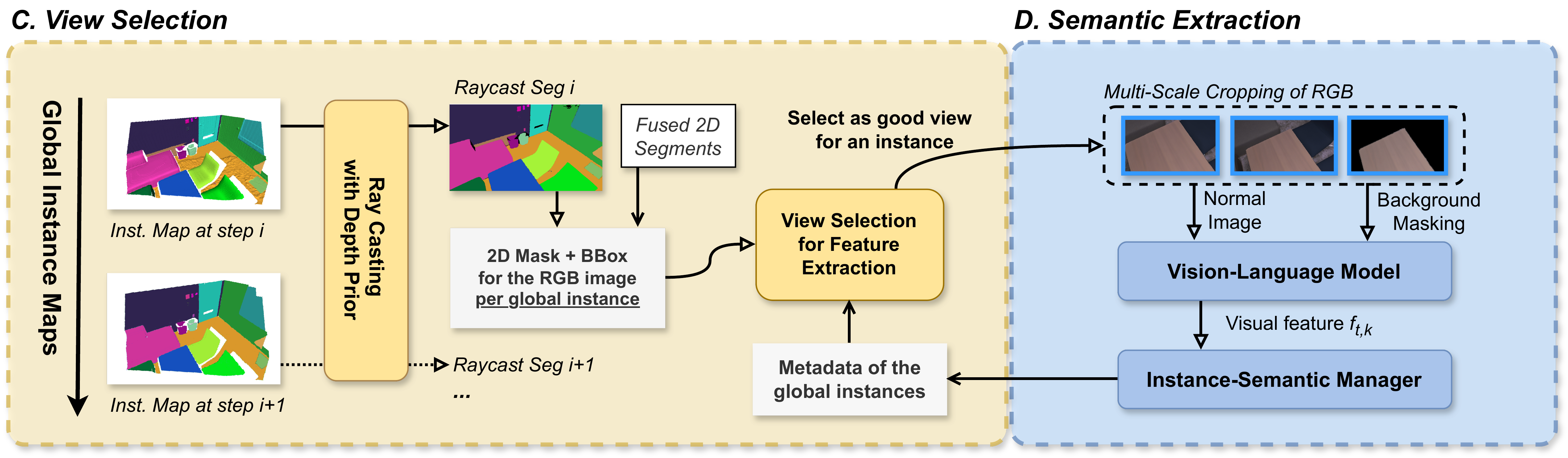}
        \caption{Incremental Semantic Features Aggregation.}
    \label{fig:pipeline2}
    \end{subfigure}
    \vspace{-1mm}
    \caption{
    \textbf{Overview} of the proposed pipeline. 
    \textbf{(a)} Class-agnostic instance map reconstruction: each RGB-D frame is segmented into entity proposals and refined with geometry-aware depth segmentation. The fused segments are lifted into 3D and incrementally integrated into a global TSDF-based instance map via super-point registration and spatial voting. 
    \textbf{(b)} Incremental semantic feature aggregation: given the global instance map, each instance is re-projected into new frames via depth-guided ray casting. A view selection module identifies informative viewpoints based on object-centric coverage. Selected views are cropped at multiple scales and masked before being passed to a Vision-Language Model (VLM). The resulting features are aggregated per instance, producing stable open-set semantic embeddings.
    }
    
    \label{fig:pipeline}
    \vspace{-3mm}
\end{figure*}

In summary, existing open-set 3D scene understanding methods either (1) store dense semantic fields in 3D, which is computationally prohibitive for incremental mapping, or (2) rely on offline or closed-set instance segmentation pipelines. 
Our work differs in two key aspects: (i) we \textit{reconstruct instances online in a class-agnostic manner}, avoiding the dependence on closed-set semantic labels for object grouping, and (ii) we introduce \textit{object-centric view selection} to efficiently aggregate VLM features, avoiding dense or redundant semantic fusion.

\section{Proposed Method}
\label{sec:method}

Given a streaming RGB-D sequence $\{(I_t, D_t), \mathbf{T}_t\}_{t = 1}^\infty$, 
where $\mathbf{T}_t \in \text{SE}(3)$ denotes the camera pose, $I_t$ and $D_t$ are the image and depth in the $t$-th frame, respectively, our goal is to incrementally construct:
(i) a \emph{class-agnostic 3D instance map}, and
(ii) a \emph{zero-shot semantic embedding} for each instance through selective multi-view feature extraction.
The key idea is to decouple instance formation from semantic inference: instances are reconstructed purely from geometric and region-consistent evidence, and semantics are assigned only when \emph{informative views} are observed.
An overview of the pipeline is shown in Fig.~\ref{fig:pipeline}. 

\subsection{Class-Agnostic Instance Map Reconstruction}
We maintain the scene in a TSDF voxel grid $\mathcal{V}$~\cite{curless1996volumetric}, following~\cite{grinvald2019voxblox++}, where each voxel $v$ stores the values for TSDF:
\[
(v_\mathrm{tsdf}, \, v_\mathrm{weight}, \, v_\ell),
\]
with $v_\ell \in \mathbb{N}$ denoting the instance identity, or $0$ if unassigned. 
Instance identities are represented via a dynamic set of 3D \emph{super-points} 
$\mathcal{S} = \{S_1, \dots, S_K\}$, where each $S_k$ corresponds to one globally consistent 3D instance and $K$ denotes the current number of instances.

\vspace{1mm}\noindent\textbf{2D Entity Segmentation and Geometric Refinement.}
For each incoming RGB-D frame, we first compute a set of class-agnostic entity proposals as follows:
\[
\{M_{t,j}\}_{j=1}^{N_t} = \mathrm{EntitySeg}(I_t),
\]
where $M_{t,j}$ is a binary mask identifying a coherent image region likely corresponding to one physical object. 
Following~\cite{qi2022high}, the segmentation emphasizes object-level boundaries while ignoring category semantics.

To improve segmentation around contacts and clutter, we compute a geometric segmentation $G_t$, in the $t$-th RGB-D frame, based on depth discontinuities~\cite{furrer2018incremental} and use it to improve the obtained 2D masks as follows:
\[
\hat{M}_{t,j} = \mathrm{MaskFusion}(M_{t,j}, G_t),
\]
where MaskFusion further split mask $G_t$ into smaller pieces according to $\{M_{t,j}\}_{j=1}^{N_t}$ to avoid under-segmentation.
This reduces over-merge errors where adjacent objects share texture or color, by using geometric discontinuities indicating boundaries not visible in appearance space.

\vspace{1mm}\noindent\textbf{3D Lifting.}
Each refined segment $\hat{M}_{t,j}$ is lifted into a 3D point cloud $P_{t,j}$ in the global coordinate frame using camera pose $\mathbf{T}_t$ and depth projection as follows:
\[
P_{t,j} = \{\mathbf{x} = \mathbf{T}_t \cdot (D_t(\mathbf{u})K^{-1}\begin{bmatrix} \mathbf{u} \\ 1 \end{bmatrix}) : \mathbf{u}\in \hat{M}_{t,j}\},
\]
where $K$ denotes the camera intrinsics, $\mathbf{u}$ is a 2D point from the region defined by mask $\hat{M}_{t,j}$, $D_t(\mathbf{u})$ is the depth observed at $\mathbf{u}$, and $\mathbf{x}$ is the 3D point.
This produces a partial surface patch corresponding to a single candidate object.

\vspace{1mm}\noindent\textbf{Instance Association via Spatial Voting.}
To determine if $P_{t,j}$ corresponds to an existing instance, we examine which instance label (\ie, index of a 3D instance) appear the most frequently in voxels covering the points in $P_{t,j}$. 

Let $V(\mathbf{x})$ denote the voxel contains $\mathbf{x}$, we define the voting of the super-point $S_k$ with label $k$ as:
\[
\Omega_{j,k} = |\{\mathbf{x}\in P_{t,j} : V(\mathbf{x})_\ell = k\}|,
\]
where $V(\mathbf{x})_\ell$ is the instance label of voxel $V(\mathbf{x})$. 
The assignment decision is the following:
\[
\small
k^* = \arg\max_k \Omega_{j,k}, \quad
\text{if } \Omega_{j,k^*} > \theta_{\mathrm{assoc}}, \text{ assign } P_{t,j} \mapsto S_{k^*};
\]
where $\theta_{\mathrm{assoc}}$ is a fixed overlapping threshold, otherwise a super-point $S_{new}$ is created with a new label.

This association depends entirely on \textit{spatial consistency}, not semantic similarity. 
Thus, unlike TSDF-based panoptic fusion pipelines~\cite{miao2024volumetric,zheng2024mapadapt}, we do not require a closed label set or hierarchical class priors.

\vspace{1mm}\noindent\textbf{Incremental TSDF Fusion and Label Stabilization.}
Once an instance with label $k^*$ is selected for $P_{t,j}$, its surface points are integrated into the TSDF grid using standard weighted fusion~\cite{newcombe2011kinectfusion}. 
Instance identity is reinforced via per-voxel label-support accumulation as follows:
\[
O_v(k^*) \leftarrow O_v(k^*) + 1 \quad \forall v = V(\mathbf{x}), \mathbf{x}\in P_{t,j}.
\]
After processing frame $t$, each voxel updates its label as:
\[
v_\ell = \arg\max_{k} O_v(k) \quad \forall v \in \mathcal{V}.
\]
This forms a temporally stable, class-agnostic 3D instance map $\mathcal{M}$ that progressively improves as the scene is observed from more viewpoints.
Moreover, over-segmentation and multi-view fusion are handled by merging spatially close super-points, introduced in the supplementary materials.

\subsection{View-Adaptive Semantic Feature Aggregation}
Once object geometry is stable, we compute open-set descriptors. 
Rather than aggregating pixel-level VLM features across all frames -- leading to noise, redundancy, and memory inefficiency -- we extract semantic features \emph{only} when a view provides novel evidence about object appearance.

\begin{figure}[t]
    \centering
    \includegraphics[width=0.98\linewidth]{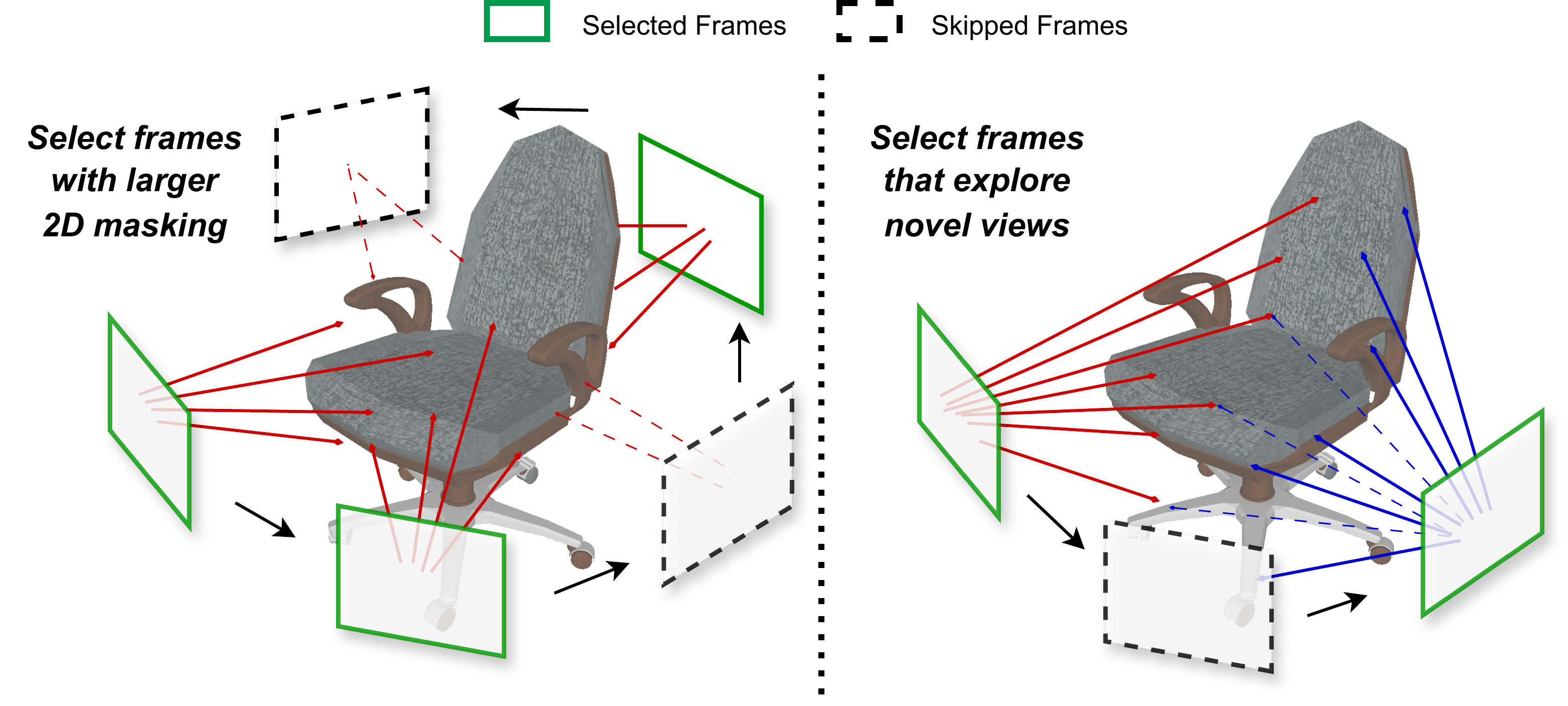}
    \caption{
    \textbf{Comparison of view selection strategies.} 
    The left illustration shows the pixel-counting strategy~\cite{takmaz2023openmask3d}, which prioritizes frames with larger object masking area, often leading to redundant front-facing views. 
    The right illustration depicts our proposed \textit{object-centric view coverage} method, which maintains a spherical map of explored viewing directions and selects frames that provide novel perspectives of the object. 
    This yields a more diverse and informative set of viewpoints for semantic feature extraction.
    }
    \label{fig:view-selection}
    \vspace{-1mm}
\end{figure}

\vspace{1mm}\noindent\textbf{Object-Centric View Coverage Representation.}
For each instance $S_k$, we maintain its \textit{view coverage} on the unit sphere. 
Let $\mathrm{Cov}_k \in \{0,1\}^{180\times240}$ denote the spherical coordinate map over bins, where if a bin contains 1, the instance has been observed from that direction.

We initialize $S_k$ a centroid $\mathbf{c}_k$ of its 3D bounding box when it gets an instance mask with area larger than a fixed threshold.
For every visible point $\mathbf{x}\in P_{t,k}$ at frame $t$, we compute its viewing direction:
\[
\mathbf{d}_{t,\mathbf{x}} = \frac{\mathbf{x}-\mathbf{c}_k}{\|\mathbf{x}-\mathbf{c}_k\|},
\]
and convert $\mathbf{d}_{t,\mathbf{x}}$ into spherical coordinates $(\mathbf{\theta}, \mathbf{\phi})$ to compute the current view coverage in the map $\mathrm{Cov}_k$.

The novelty of the current view is measured by the ratio:
\[
\eta_{t,k} = \frac{|\mathrm{BinsNewOccupied}(P_{t,k})|}{|\mathrm{BinsOccupied}(P_{t,k})|}.
\]
If $\eta_{t,k} > \theta_{\mathrm{novel}}$, the view reveals new geometric or appearance information and is selected for semantic extraction, where $\theta_{\mathrm{novel}}$ is a predefined threshold. 
After a view is selected, all bins in $\mathrm{Cov}_k$ occupied by $(\mathbf{\theta}, \mathbf{\phi})$ are set to 1.

As illustrated in Fig.~\ref{fig:view-selection}, this object-centric criterion differs fundamentally from the ``visible pixel counting'' heuristic~\cite{takmaz2023openmask3d}: while the latter repeatedly selects large, front-facing views that fail to capture full object shape, our coverage-based selection explicitly favors novel viewpoints that expand the explored surface area, yielding more diverse and informative observations.

%


\vspace{1mm}\noindent\textbf{Object-Level VLM Feature Extraction.}
For the selected frame of an instance, we extract the visual embeddings from two kinds of object crops:
1) a cropped image containing the object extent, and 
2) a masked version where background pixels are removed:
\[
\mathbf{f}_{t,k}^{(1)} = F_{\mathrm{VLM}}(I_t,P_{t,k}), \quad
\mathbf{f}_{t,k}^{(2)} = F_{\mathrm{VLM}}(I_t \odot M_{t,k}).
\]
We average these and update a running semantic feature $f_k$ per instance using visibility-aware weighting:
\[
\mathbf{f}_k \leftarrow \frac{w_\text{sum}}{w_k + w_\text{sum}} \mathbf{f}_k + \frac{w_k}{2(w_k + w_\text{sum})}  \cdot (\mathbf{f}_{t,k}^{(1)} + \mathbf{f}_{t,k}^{(2)}),
\]
where $w_k$ denotes the number of visible object pixels in frame $k$ and $w_\text{sum}$ is the cumulative pixel count over all previous observations.  
This adaptive update produces a stable, view-invariant embedding that captures semantic consistency across frames, enabling robust zero-shot reasoning and language-conditioned retrieval.

\vspace{1mm}\noindent\textbf{Why Decoupling Works.}
The decoupled design separates temporally stable instance mapping from open-vocabulary semantic inference.
Since instance mapping depends only on geometry and does not require consistent labeling, it remains stable and efficient in open-world environments. 
Semantic reasoning is then performed only when evidence is strong and informative, mitigating accumulated noise and reducing VLM calls significantly and controllably.
\section{Experiments}

\noindent
\textbf{Implementation Details.}
We adopt CropFormer~\cite{qi2022high} for 2D entity segmentation and the depth-based segmentor from~\cite{furrer2018incremental} to obtain geometric boundaries. 
Our pipeline is implemented on top of Voxblox++~\cite{grinvald2019voxblox++} using a TSDF~\cite{newcombe2011kinectfusion} representation with voxel size of 0.1m, following~\cite{miao2024volumetric} but without closed-set semantic dependencies. 
For semantic extraction, we use SigLIP~\cite{zhai2023sigmoid} as the vision-language backbone to obtain zero-shot embeddings from RGB crops. 
For evaluation, text labels are encoded with the same model, and matched to instance features via cosine similarity.
We run all methods on an Nvidia RTX3090 GPU and an Intel Core i7-12700K CPU.

\vspace{1mm} \noindent
\textbf{Datasets.}
We evaluate our method on the \textit{Replica}~\cite{straub2019replica} and \textit{ScanNet}~\cite{dai2017scannet} datasets, which provide dense RGB-D sequences with ground-truth camera poses, instance labels, and semantic annotations. 
For fair comparison, all methods are evaluated under identical input trajectories and reconstructed geometry. 
We run each sequence with 200 frames evenly sampled, as done in baselines~\cite{engelmann2024opennerf,peng2023openscene,guo2024semantic}.
For evaluation, we use the 51-class label set defined for Replica and the ScanNet200 label set for ScanNet.

\subsection{Instance Segmentation Results}

\begin{table}[t]
\centering
\resizebox{0.96\linewidth}{!}{
\begin{tabular}{l|l|c|cccc}
\hline
 & \textbf{Method} & \textbf{Online} & \textbf{mIoU} & $\mathbf{AP_{75}}$ & $\mathbf{AP_{50}}$ & $\mathbf{AP_{25}}$ \\
\hline

\multirow{4}{*}{\rotatebox{90}{Replica}}
& Mask3D        & \cross 
    & 23.1 & \underline{14.3} & \underline{31.2} & \underline{56.2} \\
& Segment3D     & \cross 
    & 29.3 & \phantom{1}6.5 & 14.5 & 24.9 \\
\cdashline{2-7}
& OVO-SLAM      & \tick 
    & \textbf{42.7} & 11.1 & 23.6 & 32.8 \\
& \textbf{Ours} & \tick 
    & \underline{36.3} & \textbf{22.0} & \textbf{50.8} & \textbf{76.7} \\
\hline

\multirow{4}{*}{\rotatebox{90}{ScanNet}}
& Mask3D        & \cross 
    & \textbf{47.6} & \textbf{16.9} & \textbf{36.1} & \textbf{47.8} \\
& Segment3D     & \cross 
    & \underline{42.0} & \phantom{1}2.5 & \phantom{1}9.3 & 16.5 \\
\cdashline{2-7}
& OVO-SLAM      & \tick 
    & 39.8 & \phantom{1}2.0 & \phantom{1}7.4 & 14.4 \\
& \textbf{Ours} & \tick 
    & 41.2 & \phantom{1}\underline{9.8} & \underline{24.0} & \underline{37.4} \\
\hline

\end{tabular}}
\caption{
\textbf{Instance segmentation performance} on Replica~\cite{straub2019replica} and ScanNet~\cite{dai2017scannet}. 
The \emph{Online} column indicates whether the method performs incremental mapping during exploration. 
We achieve comparable or better accuracy to offline approaches while consistently outperforming prior online systems, particularly at high IoU thresholds ($AP_{75}$). Mask3D was trained on ScanNet.
}
\label{tab:inst_seg_metrics}
\vspace{-3mm}
\end{table}

We first evaluate the quality of the reconstructed instance maps. 
Each reconstructed map is projected to the ground-truth mesh using a kNN-based nearest-neighbor mapping~\cite{cover1967nearest} with a fixed distance threshold, allowing per-vertex comparison with the annotated ground truth. 
We compare our approach with three state-of-the-art open-set methods: Mask3D~\cite{schult2022mask3d}, 
Segment3D~\cite{huang2024segment3d}, and OVO-SLAM~\cite{martins2024ovoslam}. 
Mask3D and Segment3D are offline volumetric networks predicting per-vertex instance masks, whereas OVO-SLAM and our method perform online, incremental mapping.

\vspace{1mm} \noindent \textbf{Metrics.}
We report mean IoU (mIoU) and instance-level average precision (AP) at IoU thresholds of 25\%, 50\%, and 75\% following~\cite{huang2024segment3d}. 
Best and second-best scores are highlighted. 
Offline methods are evaluated on the reconstructed meshes mapped to the ground truth ones.

\vspace{1mm} \noindent \textbf{Results.}
As shown in Table~\ref{tab:inst_seg_metrics}, our method achieves similar or higher instance-level precision compared to offline networks, while maintaining online operation. 
In particular, it surpasses the recent OVO-SLAM across all thresholds, with a substantial margin in $AP_{75}$, demonstrating the robustness of our class-agnostic instance tracking and fusion.
It is important to note that, while Mask3D performs the best on ScanNet, it was trained on that dataset.

\begin{figure*}[t]
    \centering
    \begin{subfigure}{0.95\linewidth}
        \includegraphics[width=\linewidth]{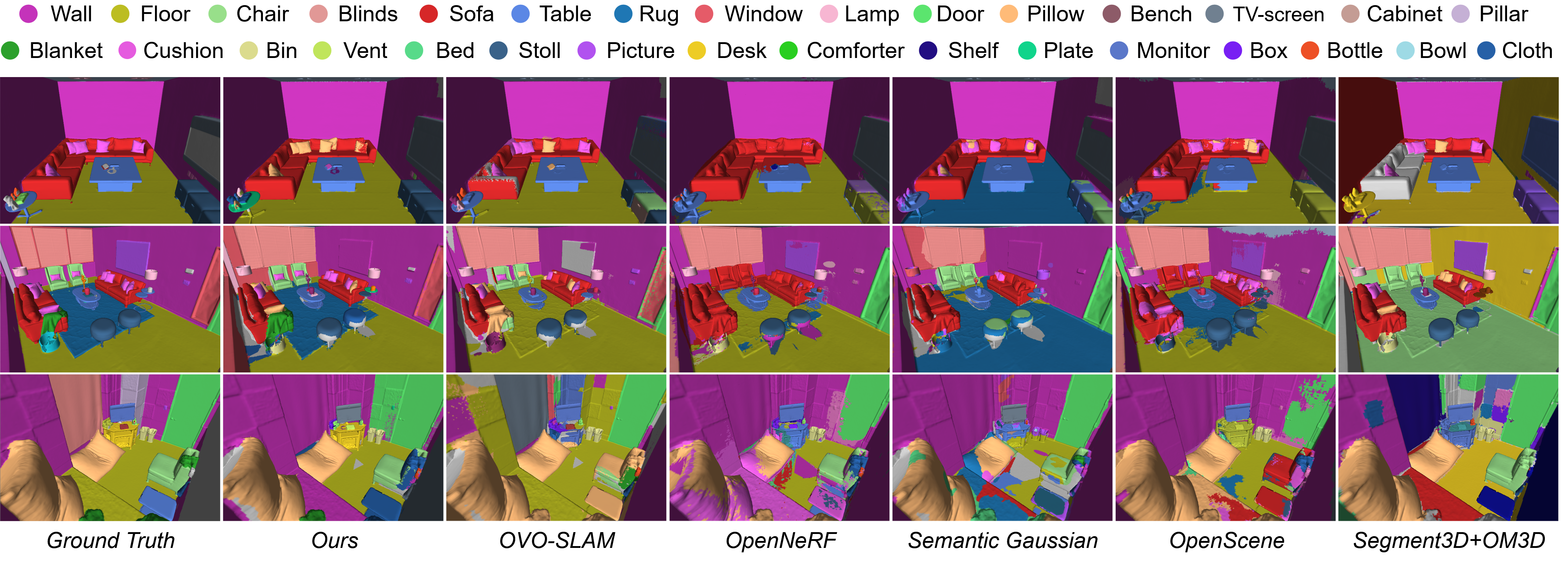}
        \caption{Qualitative comparison of semantic maps on the Replica dataset. }
    \label{fig:sem-color-visual-replica}
    \end{subfigure}
    \vspace{0.2em}
    \centering
    \begin{subfigure}{1.0\linewidth}
        \includegraphics[width=\linewidth]{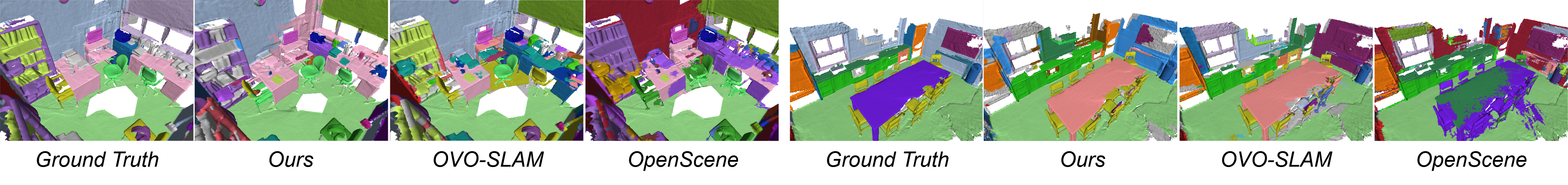}
        \caption{Qualitative comparison of semantic maps on the ScanNet dataset. }
    \label{fig:sem-color-visual-scannet}
    \end{subfigure}
    \caption{
        \textbf{Open-vocabulary 3D semantic maps} aligned to the ground-truth label sets of the respective datasets.
        We compare our method with online~\cite{martins2024ovoslam} and offline~\cite{engelmann2024opennerf,guo2024semantic,peng2023openscene,huang2024segment3d,takmaz2023openmask3d} approaches on the \textbf{\textit{Replica}} (a) and \textbf{\textit{ScanNet}} (b) datasets.
        Colors correspond to the semantic classes defined in each dataset. Gray regions indicate unobserved areas.
        OVI-MAP produces spatially coherent and semantically accurate reconstructions, maintaining sharp instance boundaries and consistent semantics throughout incremental mapping.
        Minor discrepancies in color (e.g., \textit{pillows} in (a) and the \textit{table} in (b)) arise from closed-set label mapping, where the ground truth uses alternative class names such as “cushion” or “dining table".    
    }
    \label{fig:sem-color-visual}
    \vspace{-2mm}
\end{figure*}

\subsection{Semantic Segmentation Results}

\begin{table}[t]
\centering
\resizebox{\linewidth}{!}{
\begin{tabular}{ll|c|cc|ccc}
\hline
& \textbf{Method} & \textbf{Online} & \textbf{mIoU} & \textbf{mAcc} & $\mathbf{AP_{25}}$ & $\mathbf{AP_{50}}$ & $\mathbf{AP_{all}}$ \\
\hline

\multirow{10}{*}{\rotatebox{90}{Replica}}
& Mask3D+OM3D & \cross 
    & 18.7 & 29.4 & \phantom{1}6.8 & \phantom{1}4.5 & 3.2 \\
& Segment3D+OM3D & \cross 
    & 17.3 & 30.9 & 19.9 & 15.2 & \underline{9.0} \\
& Semantic Gaussian & \cross 
    & 16.1 & 22.1 & --  & --  & -- \\
& OpenScene & \cross 
    & 19.8 & \underline{33.9} & -- & -- & -- \\
& OpenNeRF & \cross 
    & 19.2 & 30.9 & -- & -- & -- \\
& OpenFusion & \tick 
    & 14.9 & 22.4 & -- & -- & -- \\
& OVO-SLAM & \tick 
    & {24.9} & \textbf{34.0} & {28.1} & {17.5} & \textbf{9.1} \\
& \textbf{Ours} & \tick 
    & \underline{26.5} & 32.2 & \textbf{34.5} & \textbf{21.2} & 8.5 \\
\cdashline{2-8}
& OVO-SLAM (30 fps) & \tick 
    & 21.8 & 27.5 & 21.5 & 15.2 & 8.1 \\
& \textbf{Ours} (30 fps) & \tick 
    & \textbf{27.0} & {32.5} & \underline{31.8} & \underline{17.7} & {8.0} \\
\hline

\multirow{10}{*}{\rotatebox{90}{ScanNet}}
& Mask3D+OM3D & \cross 
    & \phantom{1}8.6 & 17.5 & 10.4 & \phantom{1}8.0 & 5.1 \\
& Segment3D+OM3D & \cross 
    & \phantom{1}4.5 & 13.5 & \phantom{1}5.0 & \phantom{1}3.9 & 2.3 \\
& Semantic Gaussian & \cross 
    & 10.7 & 17.8 & -- & -- & -- \\
& OpenScene & \cross 
    & 10.3 & 17.3 & -- & -- & -- \\
& OpenNeRF & \cross 
    & Fail & Fail & -- & -- & -- \\
& OpenFusion & \tick 
    & 8.3 & 11.5 & -- & -- & -- \\
& OVO-SLAM & \tick 
    & {14.6} & \textbf{27.8} & {19.4} & {12.6} & {5.5} \\
& \textbf{Ours} & \tick 
    & \textbf{17.5} & \underline{27.6} & \textbf{23.4} & \textbf{15.7} & \textbf{7.2} \\
\cdashline{2-8}
& OVO-SLAM (30 fps) & \tick 
    & 15.5 & 26.9 & 19.4 & 13.7 & 5.8 \\
& \textbf{Ours} (30 fps) & \tick 
    & \underline{16.3} & {25.4} & \underline{21.1} & \underline{14.4} & \underline{7.0} \\
\hline
\end{tabular}}
\caption{
Comparison of \textbf{open-vocabulary 3D semantic and instance segmentation} performance on {Replica}~\cite{straub2019replica} and {ScanNet}~\cite{dai2017scannet}. 
OM3D denotes {OpenMask3D}~\cite{takmaz2023openmask3d}.
The \emph{30 FPS} rows correspond to experiments where only every $n$-th frame is processed to match real-time performance. 
Our method achieves the best open-vocabulary semantic-instance accuracy among online systems, and maintains performance comparable to offline approaches even under real-time constraints.
}

\label{tab:sem_seg_metrics}
\vspace{-3mm}
\end{table}

We next evaluate the open-vocabulary semantic performance.
To further assess real-time capability, we benchmark both our {OVI-MAP} and the baseline {OVO-SLAM}~\cite{martins2024ovoslam} under a \textit{30 FPS constraint}, where only every $n$-th frame is processed for semantic extraction to run in real time.
On the Replica dataset, semantics are computed every 10th frame for our method and every 50th frame for OVO-SLAM, while on ScanNet, we process every 10th and 30th frame, respectively.
Thanks to our lightweight and view-selective semantic extraction pipeline, OVI-MAP can perform semantic updates more frequently without exceeding real-time limits.

\vspace{1mm} \noindent \textbf{Baselines.}
We compare with offline and online open-vocabulary mapping systems. 
Among the offline methods, {Mask3D+OpenMask3D (OM3D)} and {Segment3D+OM3D} combine an instance segmentation backbone with zero-shot semantic labeling: Mask3D~\cite{schult2022mask3d} and Segment3D~\cite{huang2024segment3d} generate 3D instance masks from the mesh input, while OpenMask3D~\cite{takmaz2023openmask3d} assigns open-vocabulary semantic feature to each reconstructed instance with the RGB-D input. 
We also include {Semantic Gaussian}~\cite{guo2024semantic}, {OpenScene}~\cite{peng2023openscene}, {OpenNeRF}~\cite{engelmann2024opennerf}, and OpenFusion~\cite{yamazaki2024openfusion}, which perform per-vertex open-vocabulary reasoning directly from scene representations without explicit instance segmentation. 
For online mapping, we compare to {OVO-SLAM}~\cite{martins2024ovoslam}.

\begin{figure*}[t]
    \centering
    \includegraphics[width=0.99\linewidth]{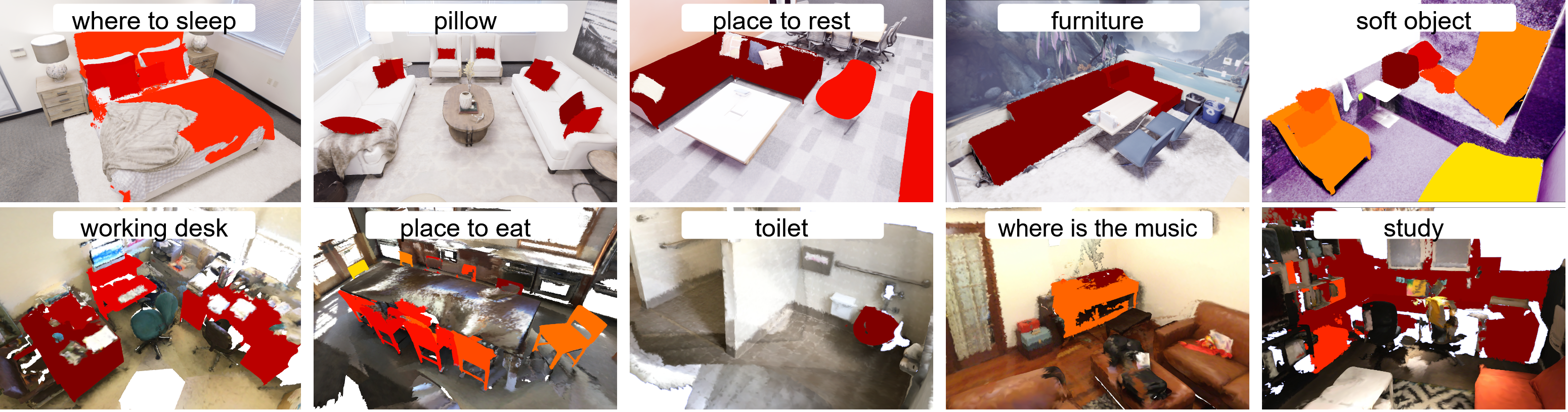}
    \caption{
        \textbf{Instance highlighting from arbitrary text queries.} 
        Given natural language prompts, our system retrieves and highlights corresponding 3D instances based on the learned vision-language embeddings. 
        The examples demonstrate zero-shot grounding of both concrete (``\textit{pillow}'', ``\textit{toilet}'') and abstract (``\textit{where to sleep}'', ``\textit{where is the music}'') concepts in reconstructed scenes.
        Darker tones indicate higher cosine similarity between an object and the query.
    }
    \vspace{-2mm}
    \label{fig:map-query}
\end{figure*}

\vspace{1mm} \noindent \textbf{Metrics.}
We report mean IoU (mIoU) and accuracy (mAcc) following~\cite{engelmann2024opennerf}, as well as instance-level precision at different IoU thresholds ($AP_{25}$, $AP_{50}$, $AP_{all}$) as done in~\cite{takmaz2023openmask3d}. 
Since the ground-truth annotations in both datasets are provided as closed-set labels, open-vocabulary features are projected onto the label set defined by the datasets, following~\cite{engelmann2024opennerf}.
The semantic feature $f_k$ of each reconstructed instance is compared against the SigLIP-encoded text embeddings of these class names, and the most similar label is assigned based on cosine similarity to enable consistent quantitative evaluation.
%

\vspace{1mm} \noindent \textbf{Results.}
As shown in Table~\ref{tab:sem_seg_metrics}, our method achieves the highest instance-level precision among the online systems and remains competitive with, or superior to, offline pipelines such as OpenScene and OpenNeRF.
On both Replica and ScanNet, our approach improves the semantic-instance $AP$ metrics by a large margin over OVO-SLAM, particularly at higher IoU thresholds ($AP_{50}$ and $AP_{25}$), demonstrating that our view-selection-based semantic aggregation yields more reliable and consistent semantics.
Under the \textit{30 FPS} real-time constraint, OVI-MAP retains most of its original performance, achieving only marginal drops in $mIoU$ and $AP$ scores, whereas OVO-SLAM exhibits a substantial degradation on Replica.
Our results outperform offline methods even in the real-time regime.

Figure~\ref{fig:sem-color-visual} presents qualitative results of the reconstructed semantic maps on Replica and ScanNet.
Our method produces spatially coherent and semantically rich reconstructions, maintaining clear object boundaries and accurate label assignments across diverse viewpoints.

\vspace{1mm} \noindent \textbf{Open Query Searching.}
Figure~\ref{fig:map-query} shows the map highlighting the most related objects in the scene with the given text queries.
These results also demonstrate the accurate instance segmentation accuracy of our method.

\begin{table}[t]
\centering
\resizebox{0.99\linewidth}{!}{
\begin{tabular}{l|l|cc|cc|c}
\hline
 & \textbf{Method} & \textbf{mIoU} & \textbf{mAcc} & $\mathbf{AP_{25}}$ & $\mathbf{AP_{50}}$ & \textbf{AQ$\downarrow$} \\
\hline

\multirow{5}{*}{\rotatebox{90}{Replica}}
& Random 8 Views 
    & \underline{23.8} & 30.4 & 31.4 & 18.6 & 50.3 \\
& Pixel Counting
    & \textbf{26.5} & \underline{31.8} & \underline{33.2} & \underline{19.8} & \underline{18.7} \\
& View Coverage (Ours) 
    & \textbf{26.5} & \textbf{32.2} & \textbf{34.5} & \textbf{21.2} & \phantom{1}\textbf{8.6} \\
\cdashline{2-7}
& \textit{GT Inst.\ + Pixel Cnt.\ }
    & 38.0 & 50.3 & 37.6 & 37.6 & 20.7 \\
& \textit{GT Inst.\ + View Cov.\ }
    & 37.6 & 48.4 & 36.2 & 36.2 & 10.4 \\
\hline

\multirow{5}{*}{\rotatebox{90}{ScanNet}}
& Random 8 Views 
    & \underline{14.7} & 22.4 & 19.2 & 13.7 & 23.0 \\
& Pixel Counting
    & \textbf{17.5} & \textbf{27.5} & \textbf{24.7} & \textbf{17.0} & \underline{15.8} \\
& View Coverage (Ours) 
    & \textbf{17.5} & \underline{27.2} & \underline{23.4} & \underline{15.7} & \phantom{1}\textbf{7.6} \\
\cdashline{2-7}
& \textit{GT Inst.\ + Pixel Cnt.\ }
    & 34.8 & 48.9 & 39.8 & 39.8 & 23.0 \\
& \textit{GT Inst.\ + View Cov.\ }
    & 30.7 & 44.8 & 34.7 & 34.7 & 12.3 \\
\hline

\end{tabular}}
\caption{
\textbf{View selection strategies} for semantic feature extraction: 
(1) random view selection, 
(2) larger visible-pixel counting similar to~\cite{takmaz2023openmask3d}, and 
(3) our proposed \textit{object-centric view coverage} method. 
\textit{AQ} denotes the average number of VLM queries per instance. 
Our approach achieves comparable zero-shot instance-level semantic precision ($AP_{25}$/$AP_{50}$) to pixel-counting while requiring substantially fewer (47\%) VLM queries.
The \textit{GT Inst.\ + Pixel Cnt.} and \textit{GT Inst.\ + View Cov.} configurations provide an upper bound using ground-truth instance masks. 
}
\vspace{-4mm}
\label{tab:ablation-view-selection}
\end{table}

\subsection{Ablation Studies}

We conduct ablation studies to analyze the effect of the proposed components, with a focus on the view-selection strategy for semantic feature aggregation and the trade-off between accuracy and computational efficiency.

\vspace{1mm}\noindent\textbf{View Selection Strategy.}
We evaluate the influence of different view selection strategies on semantic feature aggregation, focusing on the trade-off between recognition accuracy and computational efficiency. The goal is to minimize redundant queries to the vision-language model (VLM) while maintaining accurate open-vocabulary semantics.

We compare three strategies for selecting views per instance:
(1) \textit{Random} selection of eight frames per instance, serving as a naive baseline;
(2) \textit{Pixel-counting} selection, which prioritizes frames with the largest visible mask area~\cite{takmaz2023openmask3d}; and
(3) our proposed \textit{object-centric view coverage} selection, which measures how much of an object’s surface has been observed and selects frames that contribute novel surface regions.

As shown in Table~\ref{tab:ablation-view-selection}, random view selection, as expected, yields poor performance. 
Our object-centric view coverage strategy achieves comparable accuracy to pixel-counting, while using less than half the number of VLM queries (47.0\% on average). This efficiency gain arises from explicitly modeling viewpoint novelty, which avoids redundant observations while preserving semantic consistency. 

\begin{figure}[t]
    \centering
    \includegraphics[width=0.96\linewidth]{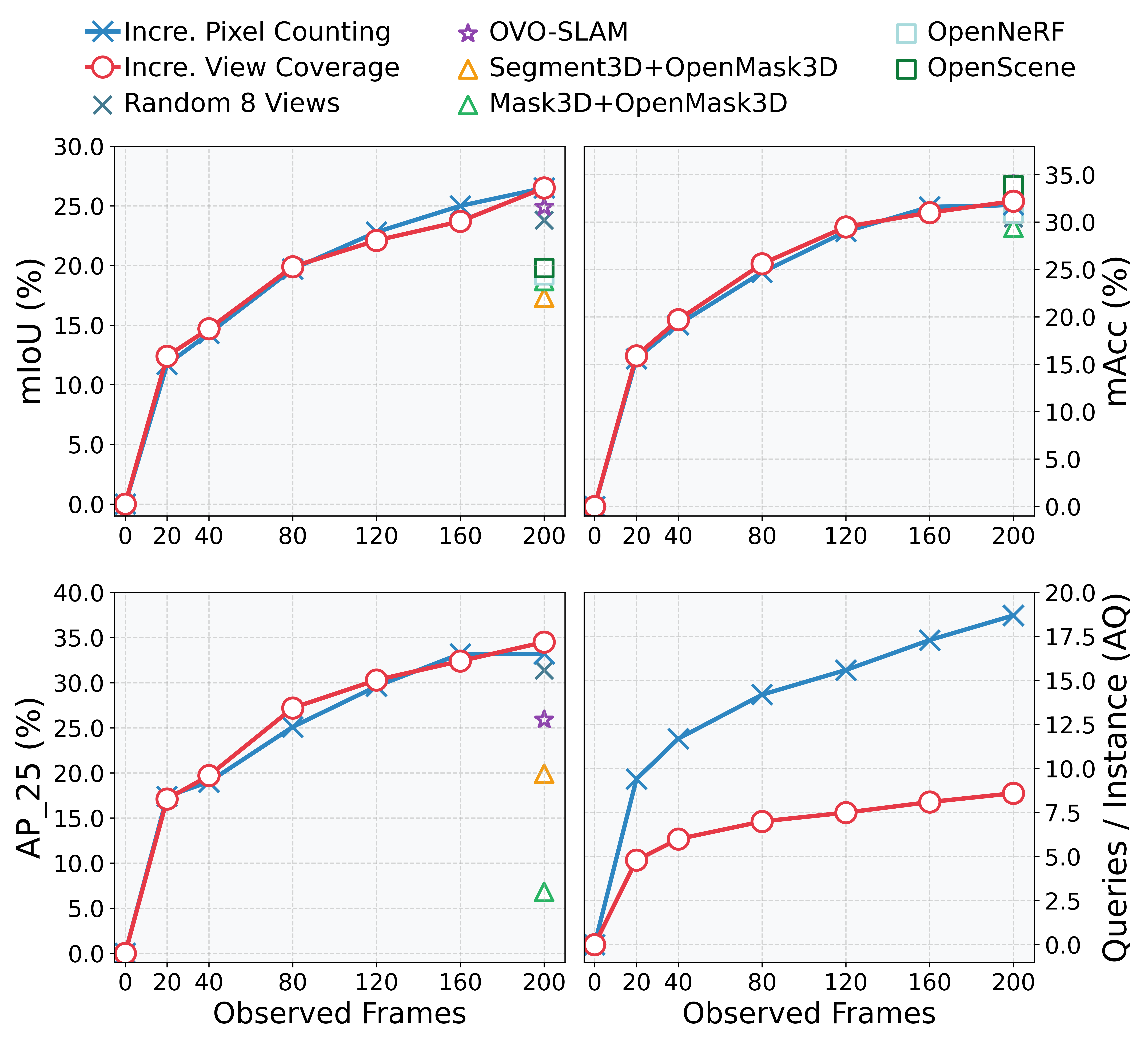}
    \caption{
        \textbf{Incremental performance} on Replica as the number of observed views increases.
        We compare our incremental semantic aggregation using the proposed \textit{view coverage} strategy against pixel-counting and other baselines. 
        Our method achieves comparable semantic accuracy (mIoU, mAcc, $AP_{25}$) while requiring significantly fewer VLM queries per instance, demonstrating efficient and scalable open-vocabulary mapping during online exploration.
    }
    \label{fig:increment-performance}
    \vspace{-3mm}
\end{figure}

\vspace{1mm}\noindent\textbf{Incremental Performance.}
Fig.~\ref{fig:increment-performance} shows how semantic accuracy evolves as more views are observed. Both our \textit{view coverage} and \textit{pixel-counting} strategies exhibit consistent improvement as additional frames are processed, demonstrating the benefits of multi-view semantic aggregation. Our view coverage approach achieves slightly better accuracy while requiring significantly fewer VLM queries per instance, as shown in the bottom-right plot.
In contrast, post-hoc semantic extraction methods -- such as OpenMask3D, Segment3D, and OpenScene -- only converge after the full sequence is processed, offering no incremental feedback during exploration. The results confirm that our coverage-based selection enables efficient, real-time semantic reasoning that scales with scene exploration, effectively bridging the gap between accuracy and online performance.

\vspace{1mm}\noindent\textbf{Influence of 2D Instance Segmentation.}
We further study how the quality of instance segmentation affects downstream semantic feature extraction. 
Table~\ref{tab:ablation-2d-instance} compares different methods for 2D instance segmentation:
(1) \textit{SAM2}, which provides high-recall but noisy over-segmentations;
(2) \textit{CropFormer}~\cite{qi2022high}, our chosen segmentor, producing compact and spatially coherent entities; and
(3) \textit{GT 2D instance masks}, serving as an oracle reference.

The results show that improving instance segmentation quality directly enhances semantic accuracy. CropFormer outperforms SAM2 in both instance- and semantic-level metrics, yielding a +5.7 improvement in $AP_{50}$ and a +6.8 gain in semantic mIoU. This confirms that cleaner and more consistent instance boundaries lead to more discriminative VLM embeddings. The use of ground-truth instance masks further boosts all metrics, demonstrating the upper bound achievable with perfect segmentation.

\vspace{1mm}\noindent\textbf{Feature Fusion.}
We analyze how semantic features from multiple selected views should be fused to form stable instance-level embeddings. Table~\ref{tab:ablation-feature-fusion} compares several strategies: simple \textit{averaging}, \textit{weighted averaging} by the number of visible object pixels, and two \textit{feature clustering} variants based on cosine similarity and $L_1$ distance.

Averaging provides a strong baseline, confirming that multi-view aggregation mitigates noise in individual predictions. Weighting by visible pixel count yields the best results across all metrics, as it naturally emphasizes clearer and better-framed object observations. Clustering-based methods offer no additional benefit and sometimes degrade performance due to overemphasis on redundant features.

These results indicate that a simple yet visibility-aware fusion is sufficient for robust zero-shot semantic reasoning, avoiding the complexity of feature clustering.

\begin{table}[t]
\centering
\resizebox{0.99\linewidth}{!}{
\begin{tabular}{l|ccc|cccc}
\hline
\multirow{2}{*}{\textbf{Method}} & \multicolumn{3}{c}{\textbf{Instance Segmentation}} & \multicolumn{4}{c}{\textbf{Semantic Segmentation}} \\
     & \textbf{mIoU} & $\mathbf{AP_{75}}$ & $\mathbf{AP_{50}}$ 
     & \textbf{mIoU} & \textbf{mAcc} & $\mathbf{AP_{25}}$ & $\mathbf{AP_{50}}$ \\
\hline
SAM2              
    & \underline{36.6} & 14.2 & 27.8 
    & 20.2 & 26.4 & 26.2 & 18.6 \\
Cropformer (Ours) 
    & 36.3 & \underline{22.0} & \underline{50.8} 
    & \textbf{26.9} & \underline{33.2} & \textbf{36.4} & \underline{22.0} \\
GT 2D Inst. Seg.  
    & \textbf{53.9} & \textbf{38.8} & \textbf{65.7} 
    & \underline{26.6} & \textbf{33.8} & \underline{36.1} & \textbf{27.3} \\
\cdashline{1-8}
\textit{GT Instances Map} 
    & 100 & 100 & 100 
    & 38.0 & 50.3 & 37.6 & 37.6 \\
\hline
\end{tabular}}
\caption{
\textbf{Influence of instance segmentation:} instance map built with 2D instance masks from (1) SAM2, (2) CropFormer, and (3) ground-truth 2D instance masks. 
All methods use the same view-selection strategy for semantic fusion. 
CropFormer yields the highest zero-shot semantic precision, while the \textit{GT Instances} row represents an upper bound with perfect instance masks.
}
\label{tab:ablation-2d-instance}
\vspace{-1mm}
\end{table}

\begin{table}[t]
\centering
\resizebox{0.98\linewidth}{!}{
\begin{tabular}{l|ccccc}
\hline
\textbf{Method} & \textbf{mIoU} & \textbf{mAcc} & $\mathbf{AP_{25}}$ & $\mathbf{AP_{50}}$ & $\mathbf{AP_{all}}$ \\
\hline
Averaging 
    & \underline{26.5} & 31.8 & \underline{33.2} & \underline{19.8} & 8.3 \\
$~$ + Weighted by Pixel Count
    & \textbf{26.9} & \textbf{33.2} & \textbf{36.4} & \textbf{22.0} & \textbf{8.6} \\
Cluster - Max. Cosine Sim.\
    & 25.1 & \underline{31.9} & 33.1 & 19.7 & \underline{8.4} \\
Cluster - Min. $L_1$ Distance
    & 24.8 & 31.2 & 32.8 & 19.5 & 8.3 \\
\hline
\end{tabular}}
\caption{
\textbf{Semantic feature fusion by} 
(1) averaging, 
(2) weighted averaging by normalized pixel counts (Ours), 
(3) selecting the feature with maximum average cosine similarity to others, and 
(4) selecting the feature with minimum average $L_1$ distance to others.
}
\label{tab:ablation-feature-fusion}
\vspace{-3mm}
\end{table}

\section{Conclusions}
\label{sec:conclusion}

Our work OVI-MAP demonstrates that open-vocabulary semantic mapping can be performed incrementally and efficiently without sacrificing accuracy. 
By combining class-agnostic instance reconstruction with language-guided feature aggregation, we show that reliable semantic reasoning is possible under online constraints. 
The proposed object-centric view coverage strategy is central to this capability, enabling informed view selection that maintains semantic consistency while substantially reducing VLM queries. 
Together, these components bridge the gap between offline open-set understanding and real-time 3D perception.

\vspace{1mm}\noindent
\textbf{Limitations.}
Our method still depends on the quality of 2D segmentation~\cite{qi2022high}, which limits performance on small or visually complex objects. 
Semantic embeddings extracted from masked RGB crops are also affected by segmentation errors and background bias. 
Finally, current vision-language models provide only weak alignment between visual and textual spaces, causing ambiguous label assignments. 
Future research will explore tighter vision-language coupling and more adaptive feature fusion to enhance robustness in cluttered, real-world environments.

\newpage
\noindent
\section*{Acknowledgments} 
This work was supported by an ETH Zurich Career Seed Award, the ETH Foundation Project 2025-FS-352, and the SNSF Advanced Grant 216260.

{
    \small
    \bibliographystyle{ieeenat_fullname}
    \bibliography{main}

@String(CVPR= {IEEE Conf. Comput. Vis. Pattern Recog.})

@String(ICCV= {Int. Conf. Comput. Vis.})

@String(ECCV= {Eur. Conf. Comput. Vis.})

@String(NIPS= {Adv. Neural Inform. Process. Syst.})

@String(TOG= {ACM Trans. Graph.})

@String(ICLR = {Int. Conf. Learn. Represent.})

@String(ICML = {Int. Conf. Mach. Learn.})

@String(IROS = {IEEE/RSJ Int. Conf. on Intell. Robots and Syst.})

@String(ICRA = {IEEE Int. Conf. on Robotics and Automation})

@String(CoRL = {Conf. Robot. Learn.})

@String(TDV = {Int. Conf. on 3D Vis.})

@String(CACM = {Commun. ACM})

@inproceedings{kirillov2023sam,
  title={Segment anything},
  author={Kirillov, Alexander and Mintun, Eric and Ravi, Nikhila and Mao, Hanzi and Rolland, Chloe and Gustafson, Laura and Xiao, Tete and Whitehead, Spencer and Berg, Alexander C and Lo, Wan-Yen and others},
  booktitle=ICCV,
  pages={4015--4026},
  year={2023}
}

@inproceedings{he2017maskrcnn,
  title={Mask r-cnn},
  author={He, Kaiming and Gkioxari, Georgia and Doll{\'a}r, Piotr and Girshick, Ross},
  booktitle=ICCV,
  pages={2961--2969},
  year={2017}
}

@inproceedings{xiong2024efficientsam,
  title={Efficientsam: Leveraged masked image pretraining for efficient segment anything},
  author={Xiong, Yunyang and Varadarajan, Bala and Wu, Lemeng and Xiang, Xiaoyu and Xiao, Fanyi and Zhu, Chenchen and Dai, Xiaoliang and Wang, Dilin and Sun, Fei and Iandola, Forrest and others},
  booktitle=CVPR,
  pages={16111--16121},
  year={2024}
}

@inproceedings{liang2023ovseg,
  title={Open-vocabulary semantic segmentation with mask-adapted clip},
  author={Liang, Feng and Wu, Bichen and Dai, Xiaoliang and Li, Kunpeng and Zhao, Yinan and Zhang, Hang and Zhang, Peizhao and Vajda, Peter and Marculescu, Diana},
  booktitle=CVPR,
  pages={7061--7070},
  year={2023}
}

@article{qi2022high,
  title={High-quality entity segmentation},
  author={Qi, Lu and Kuen, Jason and Guo, Weidong and Shen, Tiancheng and Gu, Jiuxiang and Jia, Jiaya and Lin, Zhe and Yang, Ming-Hsuan},
  journal={arXiv preprint arXiv:2211.05776},
  year={2022}
}

@article{li2020incremental,
  title={Incremental Instance-Oriented 3D Semantic Mapping via RGB-D Cameras for Unknown Indoor Scene},
  author={Li, Wei and Gu, Junhua and Chen, Benwen and Han, Jungong},
  journal={Discrete Dynamics in Nature and Society},
  volume={2020},
  number={1},
  pages={2528954},
  year={2020},
  publisher={Wiley Online Library}
}

@inproceedings{li2022lseg,
  title={Language-driven semantic segmentation},
  author={Li, Boyi and Weinberger, Kilian Q and Belongie, Serge and Koltun, Vladlen and Ranftl, Ren{\'e}},
  booktitle=ICLR,
  year={2022}
}

@inproceedings{cheng2022mask2former,
  title={Masked-attention mask transformer for universal image segmentation},
  author={Cheng, Bowen and Misra, Ishan and Schwing, Alexander G and Kirillov, Alexander and Girdhar, Rohit},
  booktitle=CVPR,
  pages={1290--1299},
  year={2022}
}

@inproceedings{ghiasi2022openseg,
  title={Scaling open-vocabulary image segmentation with image-level labels},
  author={Ghiasi, Golnaz and Gu, Xiuye and Cui, Yin and Lin, Tsung-Yi},
  booktitle=ECCV,
  pages={540--557},
  year={2022},
  organization={Springer}
}

@inproceedings{zou2023xdecoder,
  title={Generalized decoding for pixel, image, and language},
  author={Zou, Xueyan and Dou, Zi-Yi and Yang, Jianwei and Gan, Zhe and Li, Linjie and Li, Chunyuan and Dai, Xiyang and Behl, Harkirat and Wang, Jianfeng and Yuan, Lu and others},
  booktitle=CVPR,
  pages={15116--15127},
  year={2023}
}

@article{zou2023seem,
  title={Segment everything everywhere all at once},
  author={Zou, Xueyan and Yang, Jianwei and Zhang, Hao and Li, Feng and Li, Linjie and Wang, Jianfeng and Wang, Lijuan and Gao, Jianfeng and Lee, Yong Jae},
  journal=NIPS,
  volume={36},
  pages={19769--19782},
  year={2023}
}

@inproceedings{xu20243dssvlg,
  title={3d weakly supervised semantic segmentation with 2d vision-language guidance},
  author={Xu, Xiaoxu and Yuan, Yitian and Li, Jinlong and Zhang, Qiudan and Jie, Zequn and Ma, Lin and Tang, Hao and Sebe, Nicu and Wang, Xu},
  booktitle=ECCV,
  pages={87--104},
  year={2024},
  organization={Springer}
}

@inproceedings{li2024semanticsam,
  title={Segment and recognize anything at any granularity},
  author={Li, Feng and Zhang, Hao and Sun, Peize and Zou, Xueyan and Liu, Shilong and Li, Chunyuan and Yang, Jianwei and Zhang, Lei and Gao, Jianfeng},
  booktitle=ECCV,
  pages={467--484},
  year={2024},
  organization={Springer}
}

@article{ravi2024sam2,
  title={SAM 2: Segment Anything in Images and Videos},
  author={Ravi, Nikhila and Gabeur, Valentin and Hu, Yuan-Ting and Hu, Ronghang and Ryali, Chaitanya and Ma, Tengyu and Khedr, Haitham and R{\"a}dle, Roman and Rolland, Chloe and Gustafson, Laura and Mintun, Eric and Pan, Junting and Alwala, Kalyan Vasudev and Carion, Nicolas and Wu, Chao-Yuan and Girshick, Ross and Doll{\'a}r, Piotr and Feichtenhofer, Christoph},
  journal={arXiv preprint arXiv:2408.00714},
  year={2024}
}

@inproceedings{furrer2018incremental,
  title={Incremental object database: Building 3D models from multiple partial observations},
  author={Furrer, Fadri and Novkovic, Tonci and Fehr, Marius and Gawel, Abel and Grinvald, Margarita and Sattler, Torsten and Siegwart, Roland and Nieto, Juan},
  booktitle=IROS,
  pages={6835--6842},
  year={2018},
  organization={IEEE}
}

@inproceedings{takmaz2023openmask3d,
  title={Openmask3d: Open-vocabulary 3d instance segmentation},
  author={Takmaz, Ay{\c{c}}a and Fedele, Elisabetta and Sumner, Robert W and Pollefeys, Marc and Tombari, Federico and Engelmann, Francis},
  booktitle=NIPS,
  year={2023}
}

@article{takmaz2025search3d,
  title={Search3d: Hierarchical open-vocabulary 3d segmentation},
  author={Takmaz, Ayca and Delitzas, Alexandros and Sumner, Robert W and Engelmann, Francis and Wald, Johanna and Tombari, Federico},
  journal={IEEE Robotics and Automation Letters},
  year={2025},
  publisher={IEEE}
}

@inproceedings{cheng2024yoloworld,
  title={Yolo-world: Real-time open-vocabulary object detection},
  author={Cheng, Tianheng and Song, Lin and Ge, Yixiao and Liu, Wenyu and Wang, Xinggang and Shan, Ying},
  booktitle=CVPR,
  pages={16901--16911},
  year={2024}
}

@inproceedings{schult2022mask3d,
  title={Mask3d: Mask transformer for 3d semantic instance segmentation},
  author={Schult, Jonas and Engelmann, Francis and Hermans, Alexander and Litany, Or and Tang, Siyu and Leibe, Bastian},
  booktitle=ICRA,
  pages={8216--8223},
  year={2023}
}

@inproceedings{huang2024segment3d,
  title={Segment3d: Learning fine-grained class-agnostic 3d segmentation without manual labels},
  author={Huang, Rui and Peng, Songyou and Takmaz, Ayca and Tombari, Federico and Pollefeys, Marc and Song, Shiji and Huang, Gao and Engelmann, Francis},
  booktitle=ECCV,
  pages={278--295},
  year={2024},
  organization={Springer}
}

@article{yang2023sam3d,
  title={Sam3d: Segment anything in 3d scenes},
  author={Yang, Yunhan and Wu, Xiaoyang and He, Tong and Zhao, Hengshuang and Liu, Xihui},
  journal={arXiv preprint arXiv:2306.03908},
  year={2023}
}

@inproceedings{choy2019minkowski,
  title={4d spatio-temporal convnets: Minkowski convolutional neural networks},
  author={Choy, Christopher and Gwak, JunYoung and Savarese, Silvio},
  booktitle=CVPR,
  pages={3075--3084},
  year={2019}
}

@inproceedings{yin2024sai3d,
  title={Sai3d: Segment any instance in 3d scenes},
  author={Yin, Yingda and Liu, Yuzheng and Xiao, Yang and Cohen-Or, Daniel and Huang, Jingwei and Chen, Baoquan},
  booktitle=CVPR,
  pages={3292--3302},
  year={2024}
}

@inproceedings{oleynikova2017voxblox,
  title={Voxblox: Incremental 3d euclidean signed distance fields for on-board mav planning},
  author={Oleynikova, Helen and Taylor, Zachary and Fehr, Marius and Siegwart, Roland and Nieto, Juan},
  booktitle=IROS,
  pages={1366--1373},
  year={2017},
  organization={IEEE}
}

@article{grinvald2019voxblox++,
  title={Volumetric instance-aware semantic mapping and 3D object discovery},
  author={Grinvald, Margarita and Furrer, Fadri and Novkovic, Tonci and Chung, Jen Jen and Cadena, Cesar and Siegwart, Roland and Nieto, Juan},
  journal={IEEE Robotics and Automation Letters},
  volume={4},
  number={3},
  pages={3037--3044},
  year={2019},
  publisher={IEEE}
}

@article{mildenhall2021nerf,
  title={Nerf: Representing scenes as neural radiance fields for view synthesis},
  author={Mildenhall, Ben and Srinivasan, Pratul P and Tancik, Matthew and Barron, Jonathan T and Ramamoorthi, Ravi and Ng, Ren},
  journal=CACM,
  volume={65},
  number={1},
  pages={99--106},
  year={2021},
  publisher={ACM New York, NY, USA}
}

@article{kerbl20233dgs,
  title={3D Gaussian splatting for real-time radiance field rendering.},
  author={Kerbl, Bernhard and Kopanas, Georgios and Leimk{\"u}hler, Thomas and Drettakis, George},
  journal=TOG,
  volume={42},
  number={4},
  pages={139--1},
  year={2023}
}

@inproceedings{newcombe2011kinectfusion,
  title={Kinectfusion: Real-time dense surface mapping and tracking},
  author={Newcombe, Richard A and Izadi, Shahram and Hilliges, Otmar and Molyneaux, David and Kim, David and Davison, Andrew J and Kohi, Pushmeet and Shotton, Jamie and Hodges, Steve and Fitzgibbon, Andrew},
  booktitle={IEEE International Symposium on Mixed and Augmented Reality},
  pages={127--136},
  year={2011},
  organization={IEEE}
}

@inproceedings{curless1996volumetric,
  title={A volumetric method for building complex models from range images},
  author={Curless, Brian and Levoy, Marc},
  booktitle={Conference on Computer Graphics and Interactive Techniques},
  pages={303--312},
  year={1996}
}

@inproceedings{wiedmann2025dcseg,
  title={DCSEG: Decoupled 3D Open-Set Segmentation using Gaussian Splatting},
  author={Wiedmann, Luis and Wiehe, Luca and Rozenberszki, David},
  booktitle=CVPR,
  pages={5217--5226},
  year={2025}
}

@inproceedings{miao2024volumetric,
  title={Volumetric semantically consistent 3D panoptic mapping},
  author={Miao, Yang and Armeni, Iro and Pollefeys, Marc and Barath, Daniel},
  booktitle=IROS,
  pages={12924--12931},
  year={2024},
  organization={IEEE}
}

@inproceedings{qin2024langsplat,
  title={Langsplat: 3d language gaussian splatting},
  author={Qin, Minghan and Li, Wanhua and Zhou, Jiawei and Wang, Haoqian and Pfister, Hanspeter},
  booktitle=CVPR,
  pages={20051--20060},
  year={2024}
}

@inproceedings{kerr2023lerf,
  title={Lerf: Language embedded radiance fields},
  author={Kerr, Justin and Kim, Chung Min and Goldberg, Ken and Kanazawa, Angjoo and Tancik, Matthew},
  booktitle=ICCV,
  pages={19729--19739},
  year={2023}
}

@article{guo2024semantic,
  title={Semantic gaussians: Open-vocabulary scene understanding with 3d gaussian splatting},
  author={Guo, Jun and Ma, Xiaojian and Fan, Yue and Liu, Huaping and Li, Qing},
  journal={arXiv preprint arXiv:2403.15624},
  year={2024}
}

@inproceedings{jatavallabhula2023conceptfusion,
  title={Conceptfusion: Open-set multimodal 3d mapping},
  author={Jatavallabhula, Krishna Murthy and Kuwajerwala, Alihusein and Gu, Qiao and Omama, Mohd and Chen, Tao and Maalouf, Alaa and Li, Shuang and Iyer, Ganesh and Saryazdi, Soroush and Keetha, Nikhil and others},
  booktitle={Robotics: Science and Systems},
  year={2023}
}

@article{laina2025findanything,
  title={FindAnything: Open-Vocabulary and Object-Centric Mapping for Robot Exploration in Any Environment},
  author={Laina, Sebasti{\'a}n Barbas and Boche, Simon and Papatheodorou, Sotiris and Schaefer, Simon and Jung, Jaehyung and Leutenegger, Stefan},
  journal={arXiv preprint arXiv:2504.08603},
  year={2025}
}

@inproceedings{ye2024gaussiangrouping,
  title={Gaussian grouping: Segment and edit anything in 3d scenes},
  author={Ye, Mingqiao and Danelljan, Martin and Yu, Fisher and Ke, Lei},
  booktitle=ECCV,
  pages={162--179},
  year={2024},
  organization={Springer}
}

@article{yu2025panopticrecon++,
  title={Leverage cross-attention for end-to-end open-vocabulary panoptic reconstruction},
  author={Yu, Xuan and Xie, Yuxuan and Liu, Yili and Lu, Haojian and Xiong, Rong and Liao, Yiyi and Wang, Yue},
  journal={arXiv preprint arXiv:2501.01119},
  year={2025}
}

@inproceedings{zheng2024mapadapt,
  title={Map-adapt: Real-time quality-adaptive semantic 3d maps},
  author={Zheng, Jianhao and Barath, Daniel and Pollefeys, Marc and Armeni, Iro},
  booktitle=ECCV,
  pages={220--237},
  year={2024},
  organization={Springer}
}

@inproceedings{yamazaki2024openfusion,
  title={Open-fusion: Real-time open-vocabulary 3d mapping and queryable scene representation},
  author={Yamazaki, Kashu and Hanyu, Taisei and Vo, Khoa and Pham, Thang and Tran, Minh and Doretto, Gianfranco and Nguyen, Anh and Le, Ngan},
  booktitle=ICRA,
  pages={9411--9417},
  year={2024},
  organization={IEEE}
}

@article{yang2025opengsslam,
  title={OpenGS-SLAM: Open-Set Dense Semantic SLAM with 3D Gaussian Splatting for Object-Level Scene Understanding},
  author={Yang, Dianyi and Gao, Yu and Wang, Xihan and Yue, Yufeng and Yang, Yi and Fu, Mengyin},
  journal={arXiv preprint arXiv:2503.01646},
  year={2025}
}

@inproceedings{engelmann2024opennerf,
  title={OpenNeRF: open set 3D neural scene segmentation with pixel-wise features and rendered novel views},
  author={Engelmann, Francis and Manhardt, Fabian and Niemeyer, Michael and Tateno, Keisuke and Pollefeys, Marc and Tombari, Federico},
  booktitle=ICLR,
  year={2024}
}

@inproceedings{peng2023openscene,
  title={Openscene: 3d scene understanding with open vocabularies},
  author={Peng, Songyou and Genova, Kyle and Jiang, Chiyu and Tagliasacchi, Andrea and Pollefeys, Marc and Funkhouser, Thomas and others},
  booktitle=CVPR,
  pages={815--824},
  year={2023}
}

@inproceedings{zhi2021semantic-nerf,
  title={In-place scene labelling and understanding with implicit scene representation},
  author={Zhi, Shuaifeng and Laidlow, Tristan and Leutenegger, Stefan and Davison, Andrew J},
  booktitle=ICCV,
  pages={15838--15847},
  year={2021}
}

@inproceedings{yu2024panopticrecon,
  title={Panopticrecon: Leverage open-vocabulary instance segmentation for zero-shot panoptic reconstruction},
  author={Yu, Xuan and Liu, Yili and Han, Chenrui and Mao, Sitong and Zhou, Shunbo and Xiong, Rong and Liao, Yiyi and Wang, Yue},
  booktitle=IROS,
  pages={12947--12954},
  year={2024},
  organization={IEEE}
}

@inproceedings{ding2023pla,
  title={Pla: Language-driven open-vocabulary 3d scene understanding},
  author={Ding, Runyu and Yang, Jihan and Xue, Chuhui and Zhang, Wenqing and Bai, Song and Qi, Xiaojuan},
  booktitle=CVPR,
  pages={7010--7019},
  year={2023}
}

@inproceedings{li2025scenesplat,
    title={SceneSplat: Gaussian Splatting-based Scene Understanding With Vision-Language Pretraining},
    author={Li, Yue and Ma, Qi and Yang, Runyi and Li, Huapeng and Ma, Mengjiao and Ren, Bin and Popovic, Nikola and Sebe, Nicu and Konukoglu, Ender and Gevers, Theo and others},
    booktitle=ICCV,
    year={2025}
  }

@inproceedings{piekenbrinck2025opensplat3d,
  title={OpenSplat3D: Open-Vocabulary 3D Instance Segmentation using Gaussian Splatting},
  author={Piekenbrinck, Jens and Schmidt, Christian and Hermans, Alexander and Vaskevicius, Narunas and Linder, Timm and Leibe, Bastian},
  booktitle=CVPR,
  pages={5246--5255},
  year={2025}
}

@inproceedings{weder2024labelmaker,
  title={LabelMaker: automatic semantic label generation from RGB-D trajectories},
  author={Weder, Silvan and Blum, Hermann and Engelmann, Francis and Pollefeys, Marc},
  booktitle=TDV,
  pages={334--343},
  year={2024},
  organization={IEEE}
}

@inproceedings{ha2022semantic,
  title={Semantic abstraction: Open-world 3d scene understanding from 2d vision-language models},
  author={Ha, Huy and Song, Shuran},
  booktitle=CoRL,
  volume={205},
  pages={643--653},
  year={2022}
}

@article{martins2024ovoslam,
  title={OVO-SLAM: Open-Vocabulary Online Simultaneous Localization and Mapping},
  author={Martins, Tomas Berriel and Oswald, Martin R and Civera, Javier},
  journal={CoRR},
  year={2024}
}

@inproceedings{wang2025open,
  title={Open-Vocabulary Octree-Graph for 3D Scene Understanding},
  author={Wang, Zhigang and Su, Yifei and Li, Chenhui and Wang, Dong and Huang, Yan and Li, Xuelong and Zhao, Bin},
  booktitle=ICCV,
  pages={7037--7047},
  year={2025}
}

@inproceedings{deng2025openvox,
  title={Openvox: Real-time instance-level open-vocabulary probabilistic voxel representation},
  author={Deng, Yinan and Yao, Bicheng and Tang, Yihang and Zhou, Tianxing and Yang, Yi and Yue, Yufeng},
  booktitle=IROS,
  pages={1305--1311},
  year={2025},
  organization={IEEE}
}

@inproceedings{mei2025povo,
  title={Vocabulary-free 3d instance segmentation with vision-language assistant},
  author={Mei, Guofeng and Riz, Luigi and Wang, Yiming and Poiesi, Fabio},
  booktitle=TDV,
  pages={1197--1210},
  year={2025},
  organization={IEEE}
}

@inproceedings{radford2021clip,
  title={Learning transferable visual models from natural language supervision},
  author={Radford, Alec and Kim, Jong Wook and Hallacy, Chris and Ramesh, Aditya and Goh, Gabriel and Agarwal, Sandhini and Sastry, Girish and Askell, Amanda and Mishkin, Pamela and Clark, Jack and others},
  booktitle=ICML,
  pages={8748--8763},
  year={2021},
  organization={PmLR}
}

@article{shafiullah2022clipfields,
  title={Clip-fields: Weakly supervised semantic fields for robotic memory},
  author={Shafiullah, Nur Muhammad Mahi and Paxton, Chris and Pinto, Lerrel and Chintala, Soumith and Szlam, Arthur},
  journal={arXiv preprint arXiv:2210.05663},
  year={2022}
}

@inproceedings{zhai2023sigmoid,
  title={Sigmoid loss for language image pre-training},
  author={Zhai, Xiaohua and Mustafa, Basil and Kolesnikov, Alexander and Beyer, Lucas},
  booktitle=ICCV,
  pages={11975--11986},
  year={2023}
}

@article{qi2017pointnet++,
  title={Pointnet++: Deep hierarchical feature learning on point sets in a metric space},
  author={Qi, Charles Ruizhongtai and Yi, Li and Su, Hao and Guibas, Leonidas J},
  journal=NIPS,
  volume={30},
  year={2017}
}

@inproceedings{chen2022nlmap,
  title={Open-vocabulary queryable scene representations for real world planning},
  author={Chen, Boyuan and Xia, Fei and Ichter, Brian and Rao, Kanishka and Gopalakrishnan, Keerthana and Ryoo, Michael S and Stone, Austin and Kappler, Daniel},
  booktitle=ICRA,
  pages={11509--11522},
  year={2023}
}

@inproceedings{huang2022vlmap,
  title={Visual language maps for robot navigation},
  author={Huang, Chenguang and Mees, Oier and Zeng, Andy and Burgard, Wolfram},
  booktitle=ICRA,
  pages={10608--10615},
  year={2023}
}

@article{nanwani2024o3dsim,
  title={Open-set 3D semantic instance maps for vision language navigation--O3D-SIM},
  author={Nanwani, Laksh and Gupta, Kumaraditya and Mathur, Aditya and Agrawal, Swayam and Hafez, AH Abdul and Krishna, K Madhava},
  journal={Advanced Robotics},
  volume={38},
  number={19-20},
  pages={1378--1391},
  year={2024},
  publisher={Taylor \& Francis}
}

@inproceedings{dai2017scannet,
  title={Scannet: Richly-annotated 3d reconstructions of indoor scenes},
  author={Dai, Angela and Chang, Angel X and Savva, Manolis and Halber, Maciej and Funkhouser, Thomas and Nie{\ss}ner, Matthias},
  booktitle=CVPR,
  pages={5828--5839},
  year={2017}
}

@article{straub2019replica,
  title={The replica dataset: A digital replica of indoor spaces},
  author={Straub, Julian and Whelan, Thomas and Ma, Lingni and Chen, Yufan and Wijmans, Erik and Green, Simon and Engel, Jakob J and Mur-Artal, Raul and Ren, Carl and Verma, Shobhit and others},
  journal={arXiv preprint arXiv:1906.05797},
  year={2019}
}

@article{cover1967nearest,
  title={Nearest neighbor pattern classification},
  author={Cover, Thomas and Hart, Peter},
  journal={IEEE transactions on information theory},
  volume={13},
  number={1},
  pages={21--27},
  year={1967},
  publisher={IEEE}
}
}

\clearpage
\setcounter{page}{1}
\maketitlesupplementary

\section{Additional Details of the Method}

\subsection{Super-Point Merging}
In the main paper, we addressed under-segmentation in 2D instances. Over-segmentation, however, is handled by merging spatially proximate super-points. 
Here, we provide details on the merging procedure.

For all voxels containing 3D points in the point cloud $P_{t,j}$, let $\Omega_{j,k}$ denote the number of voxels assigned to the super-point $S_k$ with label $k$. 
The spatial proximity $Spa(S_a, S_b)$ measures the overlap between two super-points $S_a$ and $S_b$ by counting how many point clouds significantly intersect both:
\[
Spa(S_a, S_b) = \sum_{P_j \in \mathcal{P}} \mathbb{I}[\Omega_{j,a} > \theta_{\text{assoc}} \land \Omega_{j,b} > \theta_{\text{assoc}}],
\]
where $\mathcal{P} = \{P_{t,j}\}_{j=1}^{N_t}, \forall t \in T$ is the set of all inserted point clouds, and $\mathbb{I}[\cdot]$ is the Iverson bracket which is one of the condition inside holds and zero otherwise. 
Super-points with overlap $Spa(S_a, S_b)$ exceeding $\theta_{\text{merge}}$ are considered spatially connected, and merged into a single instance with same label.

\subsection{TSDF Map Projection via Ray-Casting}
After the incremental TSDF fusion and label stabilization with super-points voting and merging, we obtain a global TSDF-based instance map. 
We leverage this global instance map with stabilized instance labels across frames for the view-selection and feature extraction. 

We obtain the projection of the TSDF map within the current camera frame by casting ray going through each pixel to the TSDF voxels. 
Combining with the depth prior from the depth input for ray-casting, we get a globally aligned 2D instance mask that taking occlusion and multi-view consistency into account.



\section{Ablation Study on the VLM Backbone}

Table~\ref{tab:ablation-VLMs} compares several VLM backbones for semantic feature extraction, including CLIP and multiple SigLIP variants. 
Results show that the view selection strategy remains effective across different models, indicating that our method is not specific to a single VLM.
We observe that larger SigLIP variants~\cite{zhai2023sigmoid} yield consistent improvements in mIoU and $AP$ metrics, with moderate parameter growth. 
Our method uses \texttt{siglip-large-patch16-384}, which offers the best trade-off between accuracy and computational cost.


\begin{table}[t]
\centering
\resizebox{1.0\linewidth}{!}{
\begin{tabular}{l|ccccc|c}
\hline
\textbf{Method} & \textbf{mIoU} & \textbf{mAcc} & $\mathbf{AP_{25}}$ & $\mathbf{AP_{50}}$ & $\mathbf{AP_{all}}$ & \textbf{Param.}\\
\hline
clip-vit-large-patch14-336
    & 15.4 & 24.7 & 20.6 & 14.9 & 6.7 & 0.4B \\
siglip-large-patch16-384 (Ours)
    & \textbf{26.9} & {33.2} & \textbf{36.4} & {22.0} & {8.6} & 0.7B \\
siglip-so400m-patch14-384 
    & 25.2 & {33.8} & 32.4 & 19.1 & \underline{9.7} & 0.9B \\
siglip2-large-patch16-384 
    & \underline{26.7} & \underline{33.9} & \underline{35.9} & \underline{22.2} & \textbf{10.4} & 0.9B \\
siglip2-so400m-patch14-384
    & {26.4} & \textbf{34.3} & \textbf{36.4} & \textbf{23.1} & 9.4 & 1B \\
\hline
\end{tabular}}
\caption{
\textbf{Semantic feature extraction by} 
(1) ViT-L, 
(2) siglip-large (Ours), 
(3) siglip-so400m,
(4) siglip2-large, and 
(5) siglip2-so400m. The last column shows the number of parameters for each model.
}
\label{tab:ablation-VLMs}
\end{table}

\begin{table}[t]
\centering
\resizebox{1.0\linewidth}{!}{
\begin{tabular}{l|l|c|c|c}
\hline
\textbf{Method} & \textbf{Component} & \textbf{Run-Time / Frame} & \textbf{Thread} & \textbf{Skipped Frames}\\
\hline
\multirow{5}{*}{Ours} 
 & RGB Segmentation & 964.8 ms & 1 & \multirow{3}{*}{30}\\
\cdashline{2-4}
 & Depth Segmentation & \phantom{1}88.4 ms& \multirow{3}{*}{2} & \\
 & 2D-3D Association & \phantom{1}76.3 ms& & \\
\cdashline{5-5}
 & View Selection & 140.8 ms& & \multirow{2}{*}{10} \\
\cdashline{2-4}
 & Feature Extraction & 131.3 ms& 3 & \\
\hline
\multirow{3}{*}{OVO-SLAM} 
 & Segmentation & 1441.8 ms& \multirow{2}{*}{1} & \multirow{3}{*}{50}\\
 & Mapping & \phantom{1}168.8 ms& & \\
\cdashline{2-4}
 & Feature Extraction & \phantom{1}433.4 ms& 2 & \\
\hline
\end{tabular}}
\caption{
\textbf{Runtime breakdown.}  
All components run in parallel across dedicated threads. 
These steps are performed only on keyframes, meaning only every $n$-th frame is processed to match real-time performance at 30 FPS. The number of skipped frames for each component are listed in the last column.
This pipeline design keeps the overall system real-time despite potentially expensive individual modules.
}
\label{tab:runtime}
\end{table}

\section{Runtime}

Table~\ref{tab:runtime} reports the per-frame runtime of the main components, measured on an Nvidia RTX3090 GPU and an Intel Core i7-12700K CPU.
The system operates as a multi-threaded pipeline, where segmentation, 2D-3D association, view selection, and semantic feature extraction run in parallel on separate CPU and GPU threads. 
The 2D-3D association includes 3D lifting of the 2D segments, update of the super-point map, and obtaining the global instance map via ray-casting.

These instance/semantic segmentation steps are executed only on every \textit{n} frames as reported in the main paper rather than every incoming frame, which substantially reduces overall latency.

\begin{figure*}[t]
    \centering
    \begin{subfigure}{0.99\linewidth}
        \includegraphics[width=\linewidth]{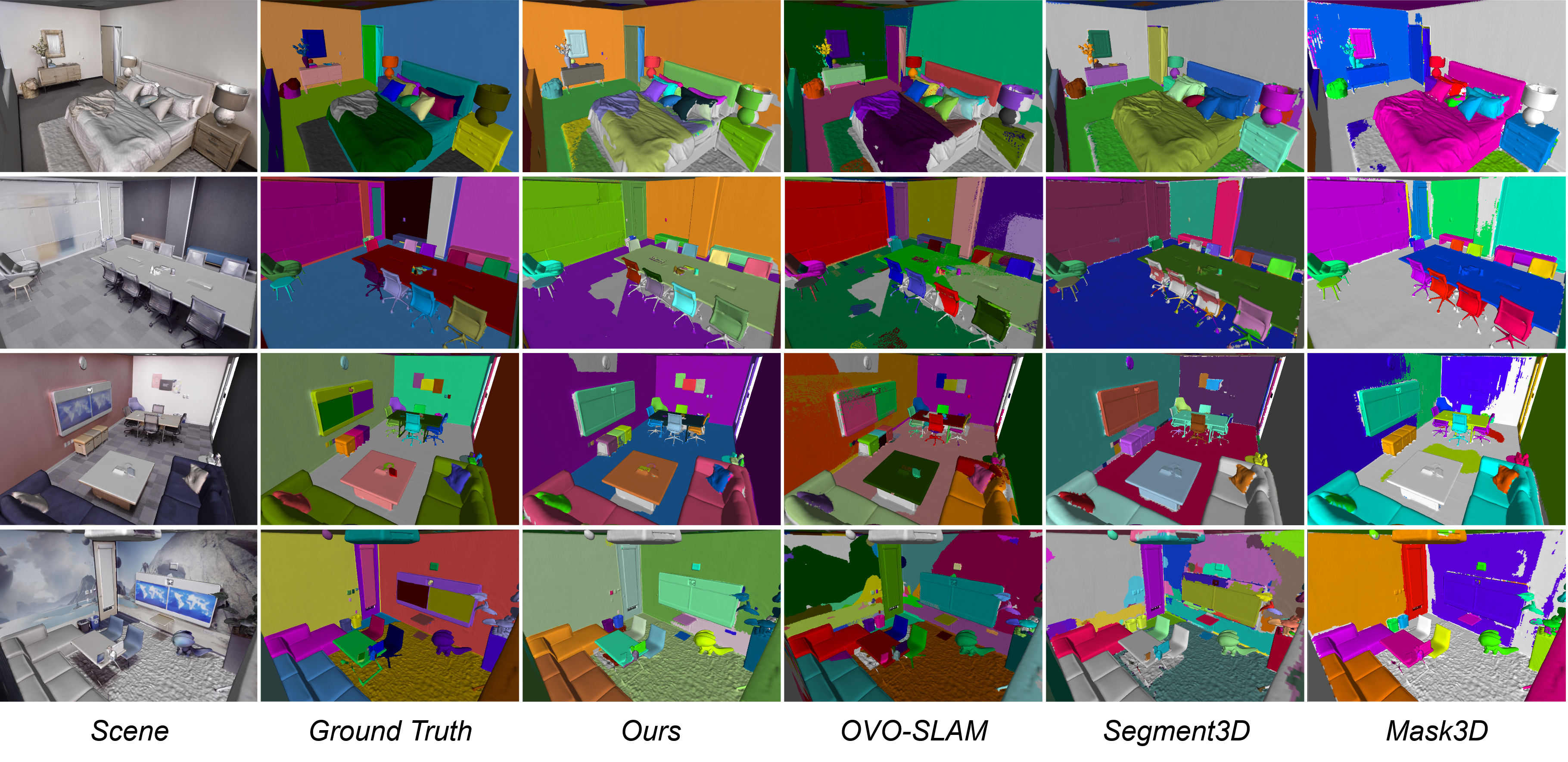}
        \caption{Qualitative comparison of instance maps on the Replica dataset. }
    \label{fig:inst-color-visual-replica}
    \vspace{1.0em}
    \end{subfigure}
    \centering
    \begin{subfigure}{0.99\linewidth}
        \includegraphics[width=\linewidth]{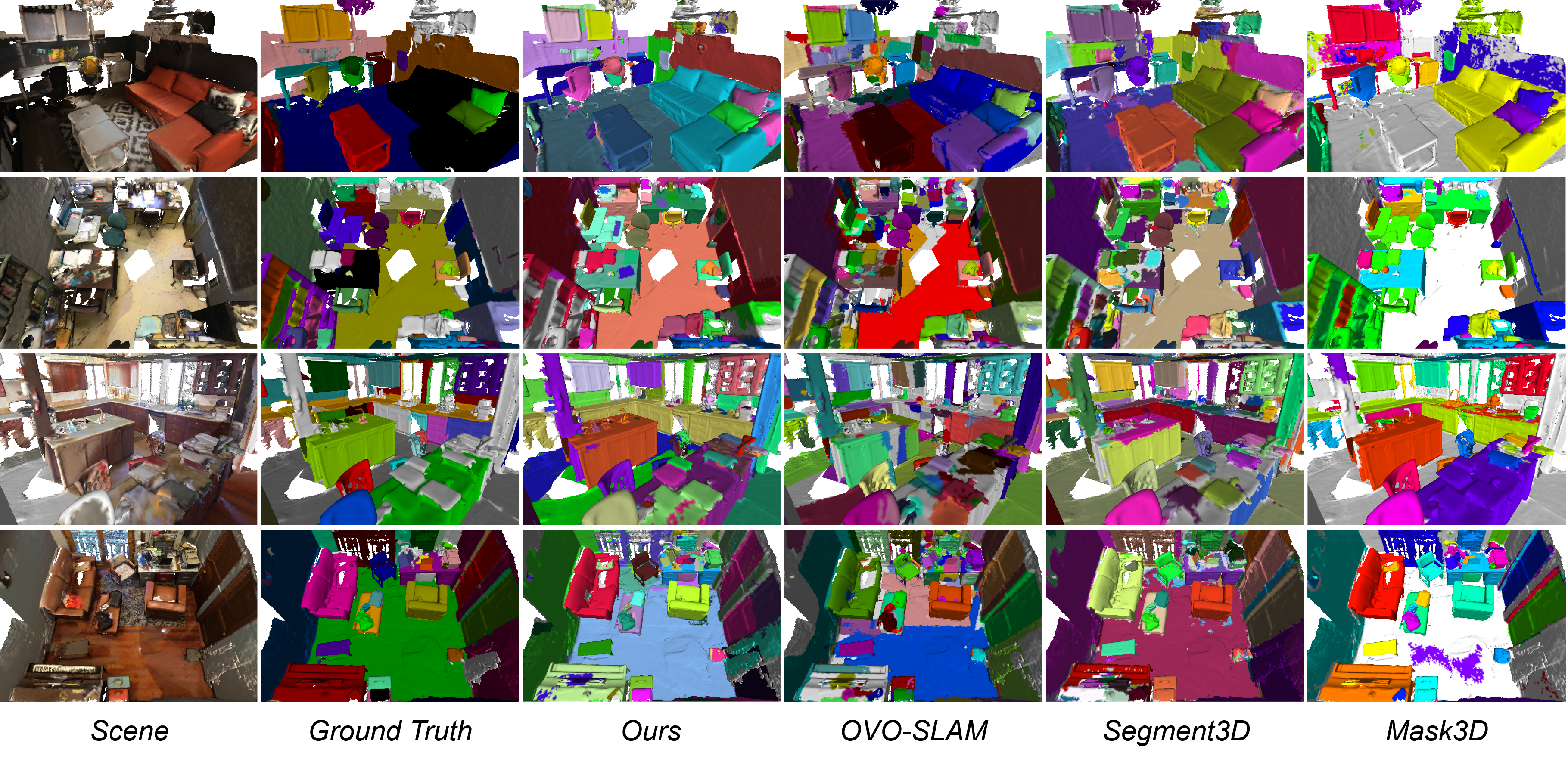}
        \caption{Qualitative comparison of instance maps on the ScanNet dataset. }
    \label{fig:inst-color-visual-scannet}
    \end{subfigure}
    \caption{
        We compare our method with online~\cite{martins2024ovoslam} and offline~\cite{huang2024segment3d,schult2022mask3d} approaches on the \textbf{\textit{Replica}} (a) and \textbf{\textit{ScanNet}} (b) datasets.
        Colors are randomly assigned for all instance maps according to the instance labels.
        Gray regions indicate unobserved areas for online methods (Ours and OVO-SLAM), and unlabeled for offline methods (Segment3D, Mask3D).
        OVI-MAP produces spatially coherent accurate reconstructions, maintaining sharp instance boundaries throughout incremental mapping.
    }
    \label{fig:inst-color-visual}
\end{figure*}

\section{Visualizations of Instance Maps}
Figure~\ref{fig:inst-color-visual} compares instance maps created by different methods. 
Online methods (Ours, OVO-SLAM) operate incrementally, while offline methods (Segment3D, Mask3D) process complete meshes. 
Colors correspond to instance IDs; gray denotes unobserved (for online methods) or unlabeled areas (for offline methods). 
Our method maintains consistent instance boundaries and achieves dense scene coverage, while offline methods leave large unlabeled regions despite full-scene access.

%

\begin{figure*}[t]
    \centering
    \begin{subfigure}{0.99\linewidth}
        \includegraphics[width=\linewidth]{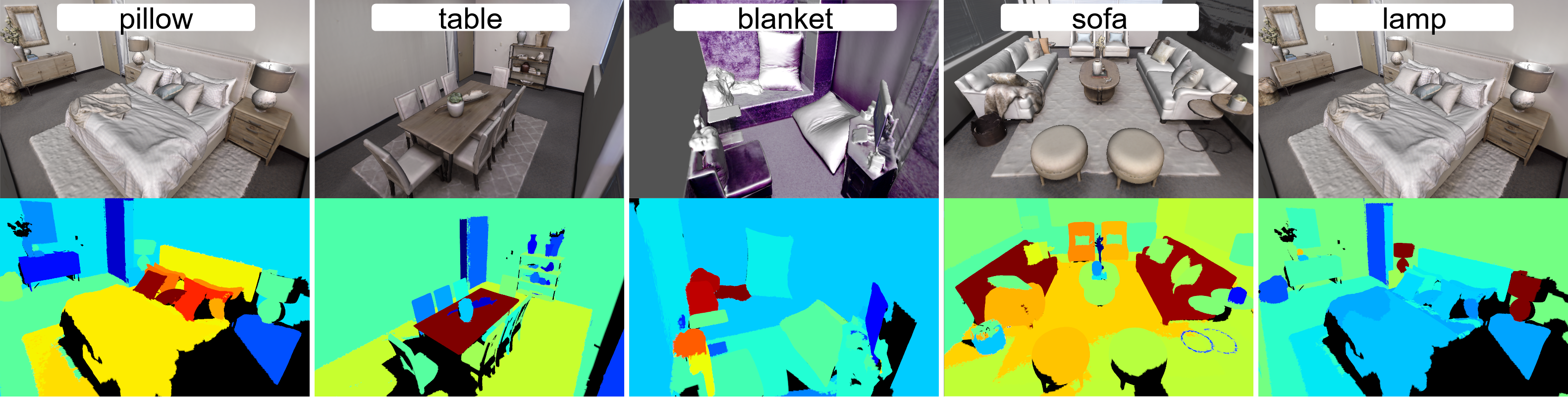}
        \caption{Heat maps for semantic querying to the scenes from the Replica dataset. }
    \label{fig:heat-map-visual-replica}
    \vspace{1.0em}
    \end{subfigure}
    \centering
    \begin{subfigure}{0.99\linewidth}
        \includegraphics[width=\linewidth]{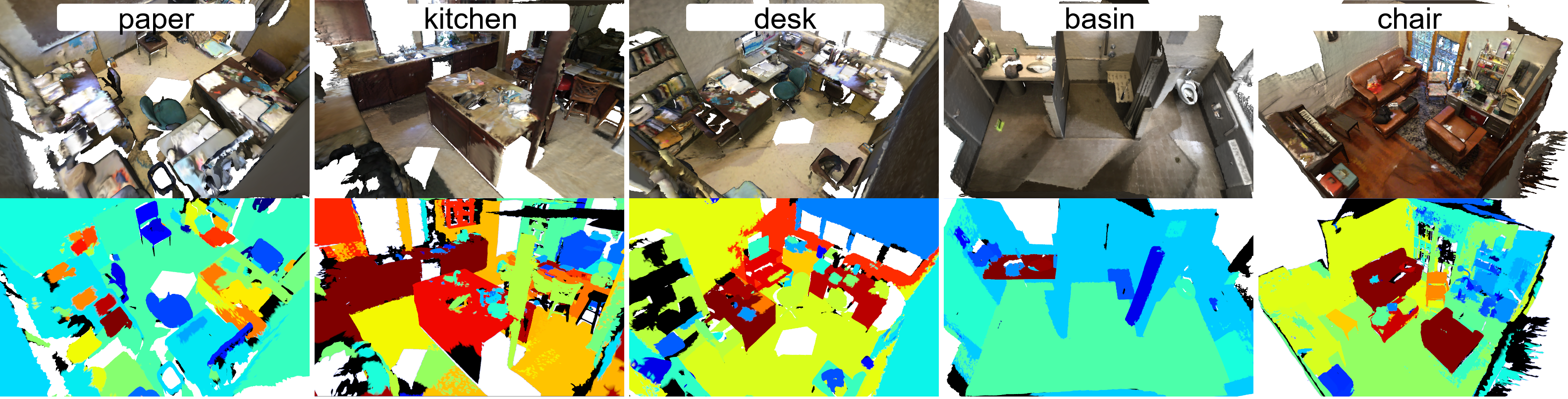}
        \caption{Heat maps for semantic querying to the scenes from the ScanNet dataset. }
    \label{fig:heat-map-visual-scannet}
    \end{subfigure}
    \caption{
        \textbf{Heat map visualizations of the semantic queries.} The color closer to red indicates the instance is more similar to the query semantic label, while the color closer to blue indicates the instance is less similar to the query semantic label. Unobserved areas are shown in black.
    }
    \label{fig:heat-map-visual}
\end{figure*}

\section{Heat Maps}

To better understand how the objects are recognized, the heat maps of the instances in the scene are shown in Fig.\ref{fig:heat-map-visual}. 
It shows us how well can the objects we query can be distinguished from the others.
We calculate the cosine similarities between the semantic features of all instances and the semantic features of the query semantic label, then map the normalized similarities to a color map.

The visualizations using heat maps further demonstrate the effectiveness of the proposed method in recognizing the objects in the scene, where the objects that match the query label are highlighted correctly, and other irrelevant objects are not highlighted as much.
This application shows the potential of the proposed method in real-world scenarios.
For example, the system can locate the objects we are looking for based on the heat map, and update the semantic map accordingly in those areas.

\section{Datasets}

We perform experiments on the Replica dataset~\cite{straub2019replica} and ScanNet~\cite{dai2017scannet}, which are widely used benchmarks for 3D scene understanding tasks. 

We use 8 scenes including \textit{'office0', 'office1', 'office2', 'office3', 'office4', 'room0', 'room1', 'room2'} for evaluation on Replica, similar to OpenNeRF~\cite{engelmann2024opennerf}.
We evaluated on 18 scenes for ScanNet, as selected by ConceptFusion~\cite{jatavallabhula2023conceptfusion}.



\end{document}